\documentclass{article}
\usepackage[utf8]{ inputenc }
\usepackage[T1]{ fontenc }
\usepackage{authblk}
\usepackage[dvipsnames]{ xcolor }
\usepackage[pdftex]{ graphicx }
\usepackage[acronym]{glossaries}
\usepackage{ geometry, enumitem }
\usepackage{booktabs}
\usepackage{multirow}
\usepackage{ algorithm, algorithmic }
\usepackage{pythonhighlight}
\usepackage{ caption, subcaption, amsmath, amssymb, dsfont, mathtools, amsthm, amsfonts, nicefrac, microtype}
\usepackage[numbers]{ natbib }
\usepackage{ hyperref }
\usepackage[capitalise, noabbrev]{ cleveref }

\title{Explainable History Distillation by Marked Temporal Point Process}

\makeatletter
\newcommand\email[2][]%
   {\newaffiltrue\let\AB@blk@and\AB@pand
      \if\relax#1\relax\def\AB@note{\AB@thenote}\else\def\AB@note{\relax}%
        \setcounter{Maxaffil}{0}\fi
      \begingroup
        \let\protect\@unexpandable@protect
        \def\thanks{\protect\thanks}\def\footnote{\protect\footnote}%
        \@temptokena=\expandafter{\AB@authors}%
        {\def\\{\protect\\\protect\Affilfont}\xdef\AB@temp{#2}}%
         \xdef\AB@authors{\the\@temptokena\AB@las\AB@au@str
         \protect\\[\affilsep]\protect\Affilfont\AB@temp}%
         \gdef\AB@las{}\gdef\AB@au@str{}%
        {\def\\{, \ignorespaces}\xdef\AB@temp{#2}}%
        \@temptokena=\expandafter{\AB@affillist}%
        \xdef\AB@affillist{\the\@temptokena \AB@affilsep
          \AB@affilnote{}\protect\Affilfont\AB@temp}%
      \endgroup
       \let\AB@affilsep\AB@affilsepx
}
\makeatother

\author[1]{Sishun Liu}
\author[1]{Ke Deng}
\author[2]{Yan Wang}
\author[1]{Xiuzhen Zhang}
\affil[1]{STEM College, RMIT University}
\email{\url{sishun.liu@student.rmit.edu}, \url{{ke.deng, xiuzhen.zhang}@rmit.edu}}
\affil[2]{School of Computing, Macquarie University}
\email{\url{yan.wang@mq.edu.au}}
\date{}

\DeclareMathOperator*{\argmax}{arg\,max}
\DeclareMathOperator*{\argmin}{arg\,min}

\hypersetup{
  colorlinks=true,
  citecolor=MidnightBlue,
  linkcolor=RoyalPurple,
  final
}

\newglossaryentry{01ip}
{
    name={\(0-1\) integer program},
    description={\(0-1\) integer program is the root of EHD.}
}

\newglossaryentry{ca}
{
    name={counterfactual analysis},
    description={Counterfactual analysis is the methodology underpinning EHD.}
}

\newglossaryentry{cif}
{
    name={conditional intensity function},
    description={One conditional intensity function defines an MTPP.}
}

\newacronym{tpp}{MTPP}{Marked Temporal Point Process}
\newacronym{mtpp}{MTPP}{Marked Temporal Point Process}
\newacronym{ehd}{EHD}{Explainable History Distillation}
\newacronym{model}{MTPP-EHD}{MTPP-based Explainable History Distiller}
\newacronym{mip}{MIP}{Mixed-Integer Programming}
\newacronym{co}{CO}{Combinatorial Optimization}
\newacronym{milp}{MILP}{Mixed-Integer Linear Programming}
\newacronym{lp}{LP}{Linear Programming}
\newacronym{st-gs}{ST-GS}{Straight-Through Gumbel-Softmax trick}
\newacronym{nll}{NLL}{Negative Log-Likelihood}
\newacronym{cfe}{CFE}{Counterfactual Explanations}
\newacronym{dppl-diff}{EHED}{Explainable History Events Distinguishment}
\newacronym{dppl-diff-metric}{DS}{Distinguishment Significance}
\newacronym{card-diff}{EMHD}{Explainable Minimal History Distillation}
\newacronym{card-diff-metric}{LMHS}{Length of Minimal Historical Subset}
\newacronym{rd}{RD}{Random Distillation}
\newacronym{gs}{GS}{Greedy Search}

\begin{document}

\newtheorem{definition}{Definition}
\newtheorem*{property}{Property}

\maketitle

\abstract{
    Explainability of machine learning models is mandatory when researchers introduce these commonly believed black boxes to real-world tasks, especially high-stakes ones. In this paper, we build a machine learning system to automatically generate explanations of happened events from history by \gls{ca} based on the \acrfull{tpp}. Specifically, we propose a new task called \acrfull{ehd}. This task requires a model to distill as few events as possible from observed history. The target is that the event distribution conditioned on left events predicts the observed future noticeably worse. We then regard distilled events as the explanation for the future. To efficiently solve \acrshort{ehd}, we rewrite the task into a \gls{01ip} and directly estimate the solution to the program by a model called \acrfull{model}. This work fills the gap between our task and existing works, which only spot the difference between factual and counterfactual worlds after applying a predefined modification to the environment. Experiment results on Retweet and StackOverflow datasets prove that \acrshort{model} significantly outperforms other  \acrshort{ehd} baselines and can reveal the rationale underpinning real-world processes.
}

\section{Introduction}
Recent time has seen much interest in applying machine learning(ML) models to high-stake real-world tasks\cite{villegas-ch_improvement_2020, rajula_comparison_2020, futoma_myth_2020, dobbelaere_machine_2021, kuleto_exploring_2021, ho_predicting_2021, vaishya_chatgpt_2023, pallathadka_impact_2023}. Because of cost and safety concerns, the explainability and accountability of these models are essential. However, most existing machine learning models are unexplainable and unaccountable black boxes. This fact forces researchers to find approaches to explain why ML models make such decisions, like showing the attention heatmap\cite{yang_ncrf_2018, chefer_generic_2021, chefer_transformer_2021}, hiring human experts to dissect reinforcement learning agents' decisions\cite{mcgrath_acquisition_2022}, or exploiting implicit causal relations, such as Granger causality\cite{xu_learning_2016, marcinkevics_interpretable_2020, zhang_cause_2020, ide_cardinality-regularized_2021, jalaldoust_causal_2022}, \Gls{ca}\cite{tan_counterfactual_2021, tsirtsis_counterfactual_2021, seedat_continuous-time_2022, noorbakhsh_counterfactual_2022}, and Wold relation\cite{etesami_variational_2021}. For users, the explanation builds their trust in existing machine learning models. For researchers, explanations help them better understand their models, enabling further performance improvements. 

The \acrfull{tpp}\cite{daley_introduction_2003} is a well-defined stochastic process that models discrete events in continuous time by a probability distribution defined over time and mark space. Learning \acrshort{tpp} by neural networks has been well investigated\cite{du_recurrent_2016, mei_neural_2017, omi_fully_2019, zhang_self-attentive_2020, zuo_transformer_2020, shchur_intensity-free_2020, mei_transformer_2021}. These algorithms enable people to train and use \acrshort{tpp} in high-stake real-world tasks like fake news modeling and mitigation\cite{farajtabar_fake_2017, zhang_vigdet_2021, zhang_counterfactual_2022} and recommendation system\cite{hosseini_recurrent_2017, cai_modeling_2018}. However, decisions made by \acrshort{mtpp} might be unreliable because causal relations between decisions and history are unknown. Recent works report that \gls{ca} could help\cite{schulam_reliable_2017, noorbakhsh_counterfactual_2022, zhang_counterfactual_2022} disclose such relations. The \gls{ca} explains model decisions by finding the smallest modification to the input features that flips the final result. For example, we assume that one piece of fake news becomes viral on X(Twitter) because two highly influential accounts retweeted it. To prove that, we remove these accounts and see if this news still goes viral in the counterfactual world. If it stopped going viral after removing both accounts but still went viral after only removing one, we would conclude that these two accounts are responsible for virality; otherwise, our assumption was wrong. 

In this paper, we investigate which parts of observed history are responsible for the future by searching for the minimal subset of observed history so that the distribution conditioned on remaining historical events significantly deviates from the observed future. We call this task the \acrfull{ehd}. As for the solution, we reform the proposed task into a \gls{01ip} and further design a machine learning algorithm called \acrfull{model} to estimate the solution of the programming problem heuristically.

In summary, the contributions of our work are:
\begin{enumerate}
    \item Following the mindset of \gls{ca}, we propose a new task, \acrfull{ehd}. It aims to explain the future by distilling the smallest subset of historical events so that the distribution conditioned on remaining historical events significantly deviates from the observed future. To the best of our knowledge, we are the first to propose this task in the \acrshort{mtpp} context.
    \item We rewrite \acrshort{ehd} into an \gls{01ip}. Then, we propose \acrfull{model}, the first machine learning algorithm that solves \acrshort{ehd} by heuristically selecting important historical events based on observed events. Extensive experiments show the superiority of \acrfull{model} by significantly outperforming existing baselines in terms of speed and the quality of distilled events.
\end{enumerate}

\section{Preliminary and Problem Statement}
\label{sec:preliminary}
\subsection{\acrlong{mtpp}} The \acrfull{mtpp} describes a random process of an event sequence \(\mathbf{x} = (x_1, x_2, \cdots, x_n)\) observed in a fixed time interval \([t_0, T]\). Each event \(x_i = (m_i, t_i)\) comprises a categorical mark \(m_i \in \mathcal{M} = \{k_1, k_2, \cdots, k_{|\mathcal{M}|}\}\) and a time \(t_i \in [t_0, T]\).

A \gls{cif} determines an \acrshort{mtpp}. Let \(\mathcal{H}_{t_l}\) denote the historical event sequence up to(include) when the most recent event happens at \(t_l\) and \(\mathcal{H}_{t-}\) denote the history up to(exclude) the current time \(t\). Given the history $\mathcal{H}_{t-}$, the \gls{cif} \(\lambda^*(m, t)\)   is the probability that event $m$ will happen at time $t$ \cite{daley_introduction_2003}\footnote{The asterisk reminds that this function conditions on history.}: 
\begin{equation}
\label{eqn:def_mtpp_intensity}
    \lambda^*(m = k_i, t) = \lim_{\Delta t \rightarrow 0}{\frac{P(m = k_i, t_{i + 1} \in (t, t+\Delta t]|\mathcal{H}_{t-})}{\Delta t}}
\end{equation}
With \(\lambda^*(m, t)\), we can define the joint probability distribution $p^*(m, t)$ of the next event where the mark is $m$ and the time is $t$.
\begin{equation}
    p^*(m, t) = \lambda^*(m, t) * \exp(-\sum_{n \in \mathcal{M}}{\int_{t_l}^{t}{\lambda^*(n, \tau)d\tau}})
\end{equation}

The negative log-likelihood on a sequence \(\mathbf{x}\) is:
\begin{equation}
\label{eqn:nll_of_mtpp}
    L = -\log p(\mathbf{x}) = - \sum_{i = 1}^{n}{\log \lambda^*(m_i, t_i)} + \sum_{j = 1}^{|\mathcal{M}|}{\int_{t_0}^{T}{\lambda^*(k_j, \tau)d\tau}}
\end{equation}
\cref{eqn:nll_of_mtpp} is widely used as the training loss of \acrshort{mtpp} models\cite{du_recurrent_2016, mei_neural_2017, omi_fully_2019, zhang_self-attentive_2020, zuo_transformer_2020, shchur_intensity-free_2020, mei_transformer_2021}. 

\subsection{\Gls{ca}}
Counterfactual analysis, also named counterfactual reasoning in some works, is one of the basic cognitive reasoning approaches\cite{tan_counterfactual_2021}. \Gls{ca} reveals the logical relations by searching for what could let things have turned out differently. For example, one bank uses a machine learning model to decide whether to accept a loan application based on the applicant's personal information. One day, Alice provided her income, age, and education attainment to this bank for a home mortgage loan. However, the bank denied her application. Alice asked: what should she do so that the bank will accept her application? The bank replied that Alice should improve her annual income by \$3000\cite{verma_counterfactual_2022}. In this case, the bank answers Alice's question by \Gls{ca} as it seeks what modification could amend the original application to be approved, which never happened in the real world. The reply reveals logical relations between Alice's application and the loan decision: the bank rejected Alice's application due to her insufficient annual income. \Gls{ca} neither requires additional knowledge about the model nor an additional explainer to find the relation between inputs and outputs. \Gls{ca} also does not change the original model, which guarantees the outcome in the future.



\subsection{\acrlong{ehd}} Suppose there is an event sequence \(\mathbf{x} = (x_1, x_2, \cdots, x_i, x_{i+1}, \cdots, x_{n-1}, x_n)\) observed in a fixed time interval \([t_0, T]\). We split \(\mathbf{x}\) at index $i$ so that the early observed events \((x_1, x_2, \cdots, x_{i})\) are considered as full history, denoted as \(\mathcal{H}_{f}\), and the recently observed events \((x_{i+1}, \cdots, x_{n-1}, x_n)\) are denoted as \(\mathbf{x}_o\). \acrfull{ehd} aims to distil the minimal subset \(\mathcal{H}_{d}\subseteq\mathcal{H}_{f}\) so that the conditional probability distribution \(p(\mathbf{x}_o | \mathcal{H}_{f})\) is significantly higher than \(p(\mathbf{x}_o | \mathcal{H}_{l})\), where \(\mathcal{H}_{l} = \mathcal{H}_{f} - \mathcal{H}_{d}\), the complement set of \(\mathcal{H}_{d}\). According to \Gls{ca}, we conclude that the distilled event sequence \(\mathcal{H}_{d}\) explains why \(\mathbf{x}_o\) happens in the factual world. Formally speaking, we expect to solve the following optimization problem:
\begin{equation}
\begin{aligned}
\label{eqn:optimization_problem}
        &\mathcal{H}_{d} = \argmin_{\mathcal{H}\subseteq\mathcal{H}_{f}}{\operatorname{card}(\mathcal{H})} \\
        &\textrm{s.t.} \quad \operatorname{dppl}(\mathcal{H}_{l}, \mathcal{H}_{f}, \mathbf{x}_o, p) = \log \operatorname{ppl}(p(\mathbf{x}_o|\mathcal{H}_{f})) - \log \operatorname{ppl}(p(\mathbf{x}_o|\mathcal{H}_{l})) < \log \epsilon; \\ 
        &\hspace{0.85cm} \mathcal{H}_{l} = \mathcal{H}_{f} - \mathcal{H}
\end{aligned}
\end{equation}
where \(\operatorname{card}(\mathcal{H})\) returns the cardinality of \(\mathcal{H}\) and \(\epsilon \in (0, 1)\) is a predefined threshold. \(\operatorname{ppl}(p(\mathbf{x}|\mathcal{H}))\) is the perplexity of conditional distribution \(p(\mathbf{x}|\mathcal{H})\) on \(\mathbf{x}\). Perplexity measures how well a distribution \(p(x)\) predicts an observed sample \(x\). A lower perplexity indicates \(p(x)\) is better at predicting \(x\). Perplexity has been used to evaluate topic modeling models\cite{hofmann_probabilistic_1999, blei_latent_2003, nallapati_multiscale_2007, hoffman_online_2010, miao_neural_2016, wang_neural_2020} and large language models(LLMs)\cite{brown_language_2020, du_glm_2022, zhang_opt_2022, zeng_glm-130b_2022}. Its definition is\cite{blei_latent_2003}:
\begin{equation}
\begin{aligned}
\label{eqn:posterior}
\operatorname{ppl}(p(\mathbf{x}_o|\mathcal{H}_{l})) &= \exp(-\frac{1}{\operatorname{card}(\mathbf{x}_o)} \log p(\mathbf{x}_o|\mathcal{H}_{l})) \\
&= \exp(-\frac{1}{\operatorname{card}(\mathbf{x}_o)} \sum_{i = 1}^{\operatorname{card}(\mathbf{x}_o)}{\log p(x_i|x_{< i}, \mathcal{H}_{l})})
\end{aligned}
\end{equation}

The difference between \(\log \operatorname{ppl}(p(\mathbf{x}_o|\mathcal{H}_{f}))\) and \(\log \operatorname{ppl}(p(\mathbf{x}_o|\mathcal{H}_{l}))\) evaluate the impact to perplexity if altering history, i.e., by distilling \(\mathcal{H}_{d}\)(removing \(\mathcal{H}_{d}\) from \(\mathcal{H}_{f}\)). According to the \Gls{ca}, the greater difference implicates that \(\mathcal{H}_{d}\) are important in explaining \(\mathbf{x}_o\). The setting of \(\epsilon\) controls to which extent the difference is high enough. We also note that various \acrshort{tpp} models can be adopted to model \(p(\mathbf{x}_o|\mathcal{H})\) in this study because the \acrshort{nll} loss in \cref{eqn:nll_of_mtpp} is the training loss of most \acrshort{tpp} models.


\section{Related Works}
\subsection{\Gls{ca}}
Machine learning researchers have introduced \Gls{ca} to explain many models, resulting in different tasks. Some researchers use \Gls{ca} to analyze how binary and multi-class classifiers make decisions and name the task \acrfull{cfe}\cite{wachter_counterfactual_2017, dhurandhar_explanations_2018, dhurandhar_model_2019, joshi_towards_2019, kanamori_dace_2020, mothilal_explaining_2020, ramakrishnan_synthesizing_2020, parmentier_optimal_2021, chen_relax_2022, xiang_realistic_2022}. The recommendation system community uses \Gls{ca} to investigate how user behaviors and item features affect recommendation results\cite{mehrotra_towards_2018, wang_information_2020, ghazimatin_prince_2020, yang_top-n_2021, tran_counterfactual_2021, wang_counterfactual_2021, xu2021learning, wang_user-controllable_2022, zhong_shap-enhanced_2022, zhang_counterfactual_2022, mu_alleviating_2022, zhang_page-link_2023}. \Gls{ca} also helps people understand how the reinforcement learning agent behaves in different environment states\cite{atrey_exploratory_2019, wang_competitive_2019, li_shapley_2021, zhou_pac_2022, ji_counterfactual_2023}. Some researchers realize that they can detect and mitigate the bias in pretrained computer-vision and language models by \Gls{ca}\cite{huang_reducing_2020, abbasnejad_counterfactual_2020, zhang_counterfactual_2020, niu_counterfactual_2021, qian_counterfactual_2021, wang_counterfactual_2022}. Recently, Noorbakhsh et al.\cite{noorbakhsh_counterfactual_2022} and Zhang et al.\cite{zhang_counterfactual_2022} introduced \gls{ca} to \acrshort{tpp} models, expecting to find how the prediction changes with handcrafted modifications to history sequences. We will give a brief introduction to some of these tasks and comparisons with \acrshort{ehd} in the following sections.


\subsubsection{\Gls{ca} on classifiers: \acrshort{cfe}}
The definition of \acrfull{cfe} involves a classifier \(f\), original input \(\mathbf{x}\), and expected output \(y\). We expect a counterfactual input \(\mathbf{x}^{\prime}\) by solving the following optimization problem:
\begin{equation}
\begin{aligned}
\label{eqn:def_cfe}
    & \mathbf{x}^{\prime} = \argmin_{\mathbf{x}^{\prime}}{d(\mathbf{x}, \mathbf{x}^{\prime})} \\
    & \mathit{s.t.} \ f(\mathbf{x}^{\prime}) = y^{\prime}
\end{aligned}
\end{equation}
where \(d(\mathbf{x}, \mathbf{x^{\prime}})\) refers to the distance between input \(\mathbf{x}\) and \(\mathbf{x^{\prime}}\). \cref{eqn:def_cfe} means the expected \(\mathbf{x}^{\prime}\) should be closer or similar to \(\mathbf{x}\) while still changes the classifier result from \(y\) to \(y^{\prime}\). Usually, the similarity between \(\mathbf{x}\) and \(\mathbf{x}^{\prime}\) means we should change as few features as possible, but sometimes it says the overall modification to \(\mathbf{x}\) should be as small as possible\cite{verma_counterfactual_2022}. \acrfull{cfe} generation is a well-investigated task\cite{wachter_counterfactual_2017, dhurandhar_explanations_2018, dhurandhar_model_2019, joshi_towards_2019, kanamori_dace_2020, mothilal_explaining_2020, ramakrishnan_synthesizing_2020, parmentier_optimal_2021, chen_relax_2022, xiang_realistic_2022}. 

Even though closely related, there are two fundamental differences between \acrshort{ehd} and \acrshort{cfe}. First, \acrshort{mtpp} is not a classification task but a regression one, so \acrshort{ehd} never involves a classifier. We must redefine the meaning of \(y\) and \(y^{\prime}\) in the regression context, inevitably proposing new problems. Second, \(\mathbf{x}\) and \(\mathbf{x}^{\prime}\) in \acrshort{cfe} are continuous, while \acrshort{ehd} optimizes the number of distilled events. It can be described by \(\mathbf{x}\)'s \(L^0\)-norm, which is a \Gls{01ip} problem. To our best knowledge, no \acrshort{cfe} approach can solve this kind of \acrshort{cfe} question. Our study tries to solve \acrshort{ehd}, thereby becoming the first \(L^0\)-sparsity, \acrshort{mtpp}-directed \Gls{ca} method.

\subsubsection{\Gls{ca} on MTPP models}
Noorbakhsh et al.\cite{noorbakhsh_counterfactual_2022} and Zhang et al.\cite{zhang_counterfactual_2022} introduced \gls{ca} to \acrshort{tpp} models. In \cite{noorbakhsh_counterfactual_2022}, the author defines \gls{ca} as a deterministic selection task among observed and rejected events generated by the thinning algorithm\cite{ogata_lewis_1981}. Changing the intensity function will deterministically accept or reject events so that the entire counterfactual process is consistent. Zhang et al. use \gls{ca} to estimate the influence of fake news engagements. They discover that users behave differently if they recently engaged in misinformation. However, the counterfactual modifications in both works are instructed by human experts, not exploited by models. On the other hand, \acrshort{ehd} expects a model that can discover the minimal subset of historical events so that the distribution conditioned on remaining historical events significantly deviates from the observed future, like what we have seen in \cite{ghazimatin_prince_2020, tan_counterfactual_2021, tran_counterfactual_2021}. Therefore, we argue that existing works can not solve our problem.


\subsubsection{\Gls{ca} on Recommendation Systems}
The literature about distilling history into a more concise sequence mainly aims at explaining the recommendation systems\cite{ghazimatin_prince_2020, tran_counterfactual_2021, xu2021learning, zhong_shap-enhanced_2022, zhang_page-link_2023}. For example, Ghazimatin et al. proposed PRINCE\cite{ghazimatin_prince_2020}, the first approach to explain recommendations concerning users' activities in Heterogeneous Information Networks(HIN). By greedily removing as few events as possible from the historical user event sequence that could replace the current recommendation with a different item, PRINCE identifies which interactions are responsible for model decisions. PRINCE heavily relies on the structure of HIN to efficiently find the solution, which limits its general use. To solve this, Tran et al. proposed ACCENT\cite{tran_counterfactual_2021}. ACCENT greedily searches for the smallest subset of history that the recommendation would change after training a new system with the subset removed.

Further, \cite{xu2021learning} uses a Variational AutoEncoder(VAE) to generate counterfactual history representations from the original one. Zhong et al.\cite{zhong_shap-enhanced_2022} discuss applying SHAP(SHapley Additive exPlanations)\cite{lundberg_unified_2017} on recommendation explanation. Zhang et al.\cite{zhang_page-link_2023} proposed PaGE-LINK, another graph-based explanation algorithm similar to PRINCE but acquires better scalability and can explain the learned GNN. However, to our best knowledge, this history distillation task has not been discussed in the \acrshort{mtpp} context. Considering these recommendation systems do not capture any temporal relations related to continuous time, these approaches can not directly apply to our task without major revisions.

\subsection{\acrlong{mip}} 
The \acrfull{mip} is a critical problem in \acrfull{co}\cite{zhang_survey_2023}. If the target function and constraints are linear, we call it \acrfull{milp}. The basic form of an \acrshort{milp} is:
\begin{equation}
\begin{aligned}
    & \min \mathbf{c}^{\top}\mathbf{x} = \sum_{i = 1}^{n}{c_ix_i} \\
    \textrm{s.t.} \quad \mathbf{A}\mathbf{x} - \mathbf{b} \geqslant 0, & \mathbf{x} \in \mathds{R}^{n}, x_i \in \mathds{Z}, i \in I, \mathbf{A} \in \mathds{R}^{m\times n}, \mathbf{b} \in \mathds{R}^{m}, c \in \mathds{R}^{n}
\end{aligned}
\end{equation}
where \(\mathbf{x}\) is a \(n\)-rank vector. \(x_i\) whose index \(i \in I\) must be integer. If every \(x_i\) in \(\mathbf{x}\) can only be 0 or 1, it becomes the \gls{01ip}, a special case of \acrshort{milp}. The original \acrshort{milp} problem is difficult to solve because the feasible region of \(\mathbf{x}\) is discrete. This means common optimization methods like convex optimization can not directly apply. For this, researchers proposed the \acrfull{lp} relaxation to relax the original problem by removing the integer restriction, as shown in \cref{eqn:tamed_MILP}.
\begin{equation}
\begin{aligned}
\label{eqn:tamed_MILP}
    & \min \mathbf{c}^{\top}\mathbf{x} = \sum_{i = 1}^{n}{c_ix_i} \\
    \textrm{s.t.} \quad \mathbf{A}\mathbf{x} - \mathbf{b} \geqslant 0, & \mathbf{x} \in \mathds{R}^{n}, \mathbf{A} \in \mathds{R}^{m\times n}, \mathbf{b} \in \mathds{R}^{m}, c \in \mathds{R}^{n}
\end{aligned}
\end{equation}
The relaxed problem is more straightforward to solve, and we might find the solution to the original \acrshort{milp} problem through the relaxed problem by branch-and-cut\cite{he_learning_2014, gasse_exact_2019, tang_reinforcement_2020}, decomposition method\cite{lange_efficient_2021}, and heuristic solutions\cite{pisinger_large_2010, khalil2017learning, chen_learning_2019, song_general_2020}. Several \acrshort{cfe} approaches employ \acrshort{milp} to find \(\mathbf{x}^{\prime}\) for linear classifiers\cite{ustun_actionable_2019, parmentier_optimal_2021, kanamori_ordered_2021} and achieve good performances. However, we notice that \acrshort{ehd}'s constraint includes \(\operatorname{dppl}(\cdot)\), the difference of two perplexities, which is highly non-linear. Such an optimization problem is not widely discussed in the literature. Moreover, direct relaxation does not work in \acrshort{ehd} because this task explicitly needs \(\mathcal{H}_{l}\) for \(\operatorname{dppl}(\cdot)\) during optimization. It indicates that all approaches mentioned above might not help solve our problem. In this paper, we will show that it is possible to estimate the solution of a \gls{01ip} by machine learning models.

\subsection{Further Related Works}
Zhang et al. report an unsupervised approach to select exogenous events from a given sequence, called TPP-Select\cite{zhang_learning_2021}. The intuition is that all observed events belong to two types: endogenous events and exogenous events. Endogenous events occur because of historical influence. On the other hand, exogenous events exist because of unknown external factors. The definition of the proposed question looks similar to \acrshort{ehd}. However, we point out that for the exogenous event selection task, the number of exogenous events is known or predefined, while in \acrshort{ehd} the algorithm should decide how many events it should remove. Moreover, TPP-Select simultaneously trains a \acrshort{mtpp} model on a new dataset with all exogenous events removed. In contrast, \acrshort{ehd} aims at extracting causal relation from a frozen \acrshort{mtpp} model. Both differences indicate that TPP-Select does not and can not solve \acrshort{ehd}.

Finally, we must clarify that \acrshort{ehd} is not Granger causality, another causal relation that people can exploit from \acrshort{mtpp} models\cite{xu_learning_2016, zhang_cause_2020, marcinkevics_interpretable_2020, zhu_vac2_2022, jalaldoust_causal_2022}. Granger causality explores mutual relations between sequences by seeing if one sequence helps forecast others. Most of These relations are among different event types. For example, Granger causality answers, "Why do the social media behaviors help the model better predict online purchasing?" On the other hand, \acrshort{ehd} model answers, "Which part of observed historical social media and online purchasing events causes the shopping heat later?". These two questions are fundamentally different.

\section{Methodology}
\label{sec:solution}
In this section, we show why and how we rewrite \acrshort{ehd} into a \Gls{01ip} and heuristically solve the \Gls{01ip} problem by our new machine learning system, \acrfull{model}. 

\subsection{Why \Gls{01ip} and the Challenges}
\label{sec:01}

Why should we rewrite \acrshort{ehd} into a \Gls{01ip}? Because the most intuitive solution to selecting a subsequence \(\mathcal{H}_{d}\) from \(\mathcal{H}_{f}\) is to introduce a binary mask tensor \(\mathbf{y} = (y_1, y_2, \cdots, y_n)\). The following will show why using \(\mathbf{y}\) inevitably rewrites \acrshort{ehd} to a \gls{01ip}. Each \(y_i \in \{0, 1\}\) tells if its corresponding event \(x_i \in \mathcal{H}_{f}\) belongs to \(\mathcal{H}_{d}\). If we believe that event \(x_i\) belongs to \(\mathcal{H}_{d}\), we let \(y_i = 1\), otherwise \(y_i = 0\). Now, the optimization target moves to the mask tensor \(\mathbf{y}\). The cardinality of \(\mathcal{H}_{d}\) changes to the number of 1s in \(\mathbf{y}\), while the constraints remain unchanged. After introducing \(\mathbf{y}\), we rewrite \cref{eqn:optimization_problem}:
\begin{equation}
\begin{aligned}
    \label{eqn:new_optimization_problem}
    \mathbf{y} &= \argmin_{\mathbf{y}}{\sum_{i = 1}^{n}{y_i}} \\
    \textrm{s.t.} \operatorname{dppl}(\mathcal{H}_{l}, \mathcal{H}_{f}, \mathbf{x}_o, p) &< \log \epsilon; \mathcal{H}_{l} = \mathcal{H}_{f} - \mathcal{H}; \forall x_i \in \mathcal{H}, y_i = 1
\end{aligned}
\end{equation}
One can realize that we have already rewritten \acrshort{ehd} into a \gls{01ip}, all starting from an intuitive solution by using mask tensor to select an \(\mathcal{H}_{d}\) from \(\mathcal{H}_{f}\). 

However, we find two difficulties that hamper training the model.
\begin{enumerate}
    \item \(\sum_{i = 1}^{n}{y_i}\) is discrete so indifferentiable. This means we can not directly optimize it. Moreover, we can not relax \(y_i\) to real numbers like what \acrshort{lp} relaxation does because we can not select \(\mathcal{H}_{d}\) and \(\mathcal{H}_{l}\) by real numbers.
    \item We need the gradient from \(\operatorname{dppl}(\mathcal{H}_{l}, \mathcal{H}_{f}, \mathbf{x}_o, p)\) to optimize \(\sum_{i = 1}^{n}{y_i}\) under the constraint. This means the conversion from \(P(y_i = C|\mathbf{x}_o, \mathcal{H}_{f})\), via \(\mathbf{y}\), to \(\mathcal{H}_{l}\) should be differentiable. However, the common practice of converting \(P(y_i = C|\mathbf{x}_o, \mathcal{H}_{f})\) to \(\mathbf{y}\) by \(\argmax(\cdot)\) and choosing \(\mathcal{H}_{l}\) from \(\mathcal{H}_{f}\) by \(\mathbf{y}\) is not differentiable.
\end{enumerate}


\subsection{\acrlong{model}}

\begin{figure}[!ht]
    \centering
    \includegraphics[width=0.65\textwidth]{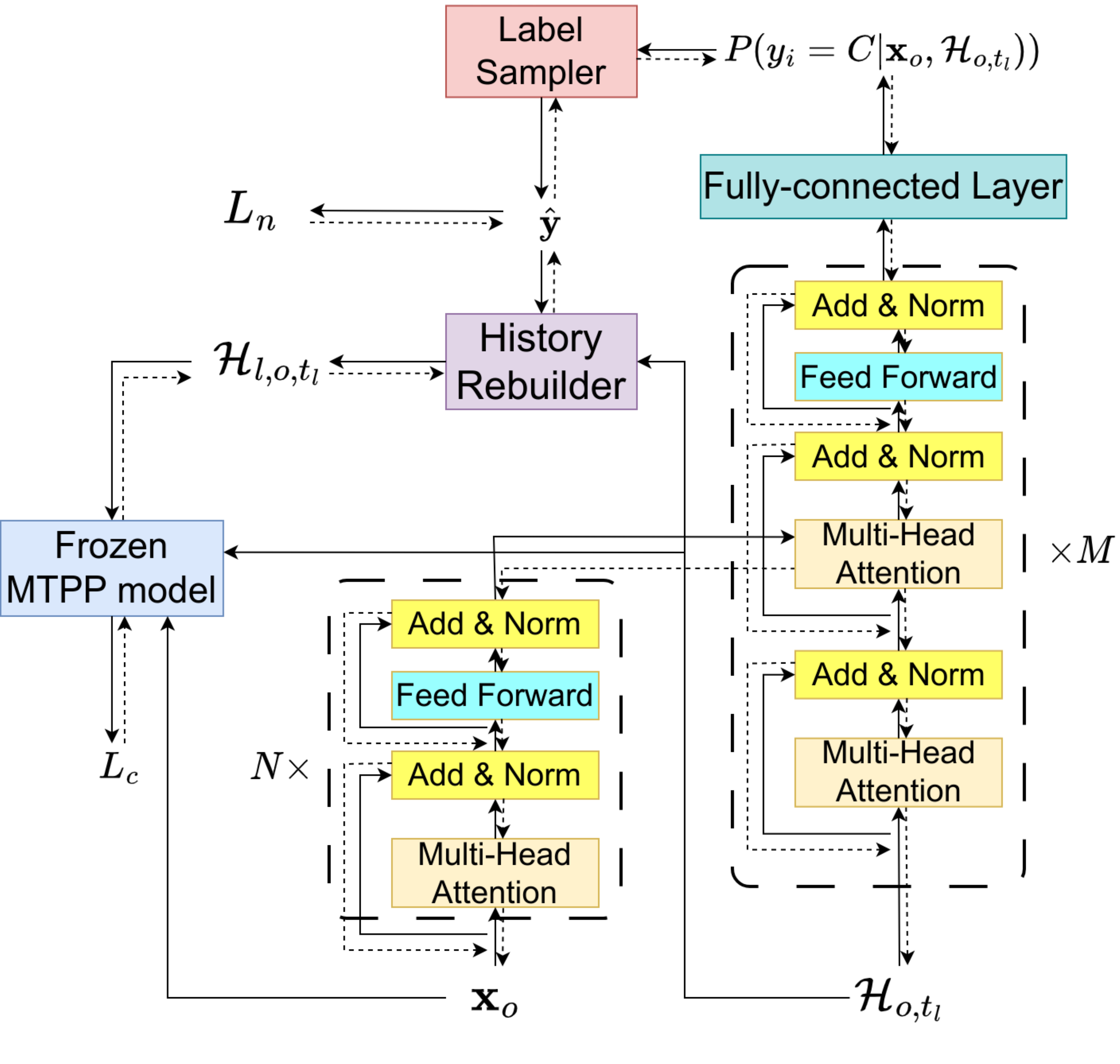}
    \caption{Architecture of \acrshort{model}. The structure of Transformer encoders comes from \cite{vaswani_attention_2017}. Black solid arrows present forward propagation paths, and black dash arrows show backpropagation paths. Two Transformer encoders(encompassed by dashed rounded rectangles) are responsible for extracting representations from \(\mathbf{x}_o\) and \(\mathcal{H}_{f}\). After mingling the representation of \(\mathbf{x}_o\) and \(\mathcal{H}_{f}\), we extract \(P(y_i = C|\mathbf{x}_o, \mathcal{H}_{f})\) by a fully-connected layer. During training, a differentiable label sampler powered by \acrshort{st-gs} trick gives \(\hat{\mathbf{y}}\)s for history rebuild. Finally, we calculate \(\operatorname{dppl}(\mathcal{H}_{l}, \mathcal{H}_{f}, \mathbf{x}_o, p)\) involving a frozen \acrshort{mtpp} model. All modules in this diagram are differentiable during training, so the gradient from \(L_c\) and \(L_n\) could tune all trainable weights.}
    \label{fig:ehd_model}
\end{figure}

Aim at solving the \gls{01ip} proposed in \cref{sec:01}, we present \acrfull{model}, a model that heuristically selects \(\mathcal{H}_{d}\) from \(\mathcal{H}_{f}\) to estimate the solution of \acrshort{ehd}. We depict the overall design of \acrshort{model} in \cref{fig:ehd_model}. \cref{alg:ehd_model_training} and \cref{alg:ehd_model_evaluation} describes how we train \acrshort{ehd} and use it to find \(\mathcal{H}_{l}\) and \(\mathcal{H}_{d}\). \acrshort{model} utilizes an encoder-decoder Transformer\cite{vaswani_attention_2017} to estimate \(P(y_i|\mathbf{x}_o, \mathcal{H}_{f})\), further optimizing the distribution according to two loss functions, \(L_n\) and \(L_c\). Below, we show the definition of \(L_n\) and \(L_c\) and why optimizing them solves \acrshort{ehd}. We also discuss how to build \(\mathcal{H}_{l}\) from \(P(y_i|\mathbf{x}_o, \mathcal{H}_{f}))\) without losing gradients.

\begin{algorithm}[tb]
\caption{Training \acrshort{model}.}
\label{alg:ehd_model_training}
\begin{algorithmic}
\STATE {\bfseries Input:} Training dataset \(D\), perplexity threshold \(\epsilon\), sample rate \(N\), training step \(tstep\);
\STATE {\bfseries Output: } The trained model \(M_{\theta}\) whose parameter is \(\theta\).

\STATE \(num\_of\_step = 0\)
\WHILE{\(num\_of\_step < tstep\)}
\STATE Fetch one observed history \(\mathcal{H}_{f}\) and observed short future \(\mathbf{x}_o\) sequence from the dataset \(D\).
\STATE \(ppl_f = \operatorname{ppl}(p_{model}(\mathbf{x}_o|\mathcal{H}_{f}))\)
\STATE \(P(y_i = C|\mathcal{H}_{f}, \mathbf{x}_o) = M_{\theta}(\mathcal{H}_{f}, \mathbf{x}_o)\)
\STATE Sample \(N\) \(\hat{\mathbf{y}}\)s from \(P(y_i = C|\mathcal{H}_{f}, \mathbf{x}_o)\) by the ST-GS trick.
\STATE Build \(\mathcal{H}_{i,l,o,t_l}\) from every sampled \(\hat{\mathbf{y}}\). We totally collect \(N\) \(\mathcal{H}_{i,l,o,t_l}\)s.
\STATE Calculate \(L = L_c + L_n\)
\STATE Update \(\theta \leftarrow \theta - learning\_rate * \nabla_{\theta}L\)
\STATE \(num\_of\_step = num\_of\_step + 1\)
\ENDWHILE
\RETURN \(M_{\theta}\)

\end{algorithmic}
\end{algorithm}

\begin{algorithm}[tb]
\caption{Using \acrshort{model} to decide \(\mathcal{H}_{d}\) given \(\mathcal{H}_{f}\) and \(\mathbf{x}_o\).}
\label{alg:ehd_model_evaluation}
\begin{algorithmic}
\STATE {\bfseries Input:} Trained model \(M_{\theta}\), The observed history \(\mathcal{H}_{f}\), observed short future \(\mathbf{x}_o\)
\STATE {\bfseries Output:} \(\mathcal{H}_{l}\) and \(\mathcal{H}_{d}\)

\STATE \(P(y_i = C|\mathcal{H}_{f}, \mathbf{x}_o) = M_{\theta}(\mathcal{H}_{f}, \mathbf{x}_o)\)
\STATE \(\mathbf{y} = \argmax(P(y_i = C|\mathcal{H}_{f}, \mathbf{x}_o))\)
\STATE Build \(\mathcal{H}_{l}\) and \(\mathcal{H}_{d}\) according to \(\mathbf{y}\). 
\RETURN \(\mathcal{H}_{l}\), \(\mathcal{H}_{d}\)

\end{algorithmic}
\end{algorithm}

The core issue that hinders solving \acrshort{ehd} is how to differentially generate \(\mathcal{H}_{l}\) and \(\mathcal{H}_{d}\) from \(P(y_i = C|\mathbf{x}_o, \mathcal{H}_{f})\) via \(\mathbf{y}\). \(\mathbf{y}\) is a discrete mask tensor containing only 0 and 1, telling events belonging to whether \(\mathcal{H}_{l}\) or \(\mathcal{H}_{d}\). Discrete tensors can not have gradients. To fix this, we use the \acrfull{st-gs}\cite{maddison_concrete_2016, bengio_estimating_2013} to sample a differentiable yet continuous mask \(\hat{\mathbf{y}} = \{\hat{y_1}, \hat{y_2}, \cdots, \hat{y_n}\}\). \acrshort{st-gs} trick samples differentiable \(\hat{y_i}\)s following the categorical distribution \(P(y_i = C|\mathbf{x}_o, \mathcal{H}_{f})\) using \cref{eqn:st_gs}.
\begin{equation}
\label{eqn:st_gs}
 \hat{y_i} = \frac{\exp((\log P(y_i = C|\mathbf{x}_o, \mathcal{H}_{f}) + g_C) / \tau)}{\sum_{j \in \{0, 1\}}{\exp(\log P(y_i = j|\mathbf{x}_o, \mathcal{H}_{f}) + g_j) / \tau)}}
\end{equation}
where \(\tau\) refers to the temperature and \(g_i\) are i.i.d. samples from the standard Gumbel distribution. The Straight-Through trick enables us to set \(\tau = 0\) during forwardpropagation and estimates its gradient by a proxy function, namely the derivative of \cref{eqn:st_gs} with \(\tau > 0\). This trick allows \(\hat{\mathbf{y}}\) only to contain 0 and 1 and still send gradients to \(P(y_i = C|\mathbf{x}_o, \mathcal{H}_{f})\). 

With \(\hat{\mathbf{y}}\), we can discuss how we solve \acrshort{ehd}, namely the definition of \(L_c\) and \(L_n\) as \(\hat{\mathbf{y}}_i\) is now differentiable. 

\textbf{Loss \(L_c\) for enforcing the constraints} Inspired by \cite{mothilal_explaining_2020, tan_counterfactual_2021}, we relax the constraint as a hinge loss:
\begin{equation}
\label{eqn:hinge_loss}
\begin{aligned}
    L(\hat{\mathbf{y}}, \mathbf{x}_o, \mathcal{H}_{f}, p_{model}, \epsilon) = \max(\operatorname{dppl}(\mathcal{H}_{l}, \mathcal{H}_{f}, \mathbf{x}_o, p_{model}) - \log \epsilon, 0)
\end{aligned}
\end{equation}
Only one sampled \(\hat{\mathbf{y}}\) may introduce highly inconsistent gradients during training because the value of \(\hat{\mathbf{y}}\) changes in each sample. We mitigate this issue by sampling \(N\) \(\hat{\mathbf{y}}_i\)s from \(P(y|\mathbf{x}_{o}, \mathcal{H}_{f})\) next calculating the expectation of \(L(\mathbf{y}, \mathbf{x}_o, \mathcal{H}_{f}, p_{model}, \epsilon)\) over all \(\mathcal{H}_{i,l,o,t_l}\)s rebuilt from \(\hat{\mathbf{y}}_i\)s, which is \(L_c\):


\begin{equation}
\begin{aligned}
    \label{eqn:L_c}
    L_c & = \mathds{E}_{\hat{\mathbf{y}}\sim P(y|\mathbf{x}_o, \mathcal{H}_{f})}{L(\hat{\mathbf{y}}, \mathbf{x}_o, \mathcal{H}_{f}, p_{model}, \epsilon)} \\
    & = \mathds{E}_{\hat{\mathbf{y}}\sim P(y|\mathbf{x}_o, \mathcal{H}_{f})}[\max(\operatorname{dppl}(\mathcal{H}_{l}, \mathcal{H}_{f}, \mathbf{x}_o, p_{model}) - \log \epsilon, 0)] \\
    & \approx \frac{1}{N}\sum_{i = 1}^{N}{\max(\operatorname{dppl}(\mathcal{H}_{i, l, o, t_l}, \mathcal{H}_{f}, \mathbf{x}_o, p_{model}) - \log \epsilon, 0)}
\end{aligned}
\end{equation}
where \(\mathcal{H}_{i,l,o,t_l}\) refers to a specific \(\mathcal{H}_{l}\) rebuilt from the \(i\)th sampled \(\hat{\mathbf{y}}_i\). After differentiably rebuilding the new history sequence \(\mathcal{H}_{l}\) using \(\hat{\mathbf{y}}\), we utilize a trained \acrshort{tpp} model to estimate the perplexity. Since plenty of existing \acrshort{tpp} models are trained using the \acrfull{nll} loss as shown in \cref{eqn:nll_of_mtpp}, we can plug in one of these models into \acrshort{ehd}. The only requirement is that the \acrshort{mtpp} model is differentiable so \acrshort{model} could obtain the gradient \(\nabla_{\hat{\mathbf{y}_i}}{L_c}\) to enable training. In this paper, the trained \acrshort{mtpp} model is FullyNN\cite{omi_fully_2019}. 

\textbf{Loss \(L_n\) for optimizing the size of \(\mathcal{H}_{d}\)} Following the definition of \acrshort{ehd}, the size of \(\mathcal{H}_{d}\) is the number of non-zero values in \(\mathbf{y}\), or \(L^0\) norm of \(\mathbf{y}\). In common practice, the number of non-zero values in a vector \(\mathbf{v}\), known as the \(L^0\) norm, is indifferentiable. As a workaround, people optimize \(\mathbf{v}\)'s differentiable \(L^1\) norm\cite{tan_counterfactual_2021}. However, optimizing the \(L^1\) of a vector \(\mathbf{v}\in \mathds{R}^{d}\) has limited effects on optimizing \(L^0\) because there is no consistent monotonicity relation between \(L^0\) and \(L^1\). \(L^0\) could decrease, stay unchanged, or even increase when \(L^1\) decreases. But, this statement is not true specifically for \(\hat{\mathbf{y}}\). Why? \acrshort{st-gs} ensures that \(\hat{\mathbf{y}}\) only contains 0 and 1, so for \(\hat{\mathbf{y}}\), \(L^0\) is always equal to \(L^1\). This means optimizing \(\hat{\mathbf{y}}\)'s \(L^1\) is equivalent to optimizing \(\hat{\mathbf{y}}\)'s \(L^0\). \(L^1\) is differentiable, so we can use it to optimize \(\operatorname{card}(\mathcal{H}_{d})\):
\begin{equation}
\label{eqn:l_n}
    L_n = \frac{L^0(\hat{\mathbf{y}})}{\operatorname{card}(\hat{\mathbf{y}})} = \frac{L^1(\hat{\mathbf{y}})}{\operatorname{card}(\hat{\mathbf{y}})}
\end{equation}
The \(L^1\) norm tends to be much larger than \(L_c\). We divide the \(L^1\) by \(\hat{\mathbf{y}}\)'s cardinality to normalize it. 

Now, we can express the training loss of \acrshort{model}. The training loss \(L\), as shown in \cref{eqn:loss}, is the sum of \(L_n\) and \(L_c\). We use a hyperparameter \(\alpha\) to balance the trade-off between the number of distilled events and the probability drop.
\begin{equation}
\label{eqn:loss}
    L = \alpha L_n + L_c
\end{equation}
Other training model details, including hyperparameters and technical information about the history rebuilder in \cref{fig:ehd_model}, are available in \cref{app:model_training}.

\section{Experiment settings}
In this section, we express the \acrshort{ehd} baseline models, \acrshort{ehd} tasks and corresponding evaluation metrics, and used datasets. 

\label{sec:settings}
\subsection{Baseline Models}
To our best knowledge, we are the first to propose \acrshort{ehd} in the \acrshort{mtpp} context. This means we do not have appropriate baselines to compare with. Brute force is never a plausible approach to \acrshort{ehd} because \gls{01ip} is NP-hard\cite{karp_reducibility_1972}. After a deep investigation of existing works regarding explaining recommendation systems, we notice that most of them are built upon \acrfull{gs}\cite{ghazimatin_prince_2020, tran_counterfactual_2021, zhong_shap-enhanced_2022}. \textbf{\acrfull{gs}} can solve \acrshort{ehd} by searching for the next event that could decrease \(\operatorname{dppl}(\mathcal{H}_{l}, \mathcal{H}_{f}, \mathbf{x}_o, p)\) most until it reaches a desired target. We also set up the \textbf{\acrfull{rd}} baseline, which randomly moves events from \(\mathcal{H}_{f}\) to \(\mathcal{H}_{d}\), to show the difficulty of \acrshort{ehd}. Detailed information about these baselines is in \cref{app:baselines}.

\subsection{Evaluation Tasks and Metrics}
We evaluate \acrshort{ehd} performance of all approaches by two novel tasks paired with new metrics. First task requires the difference between \(\operatorname{dppl}(\mathcal{H}_{d}, \mathcal{H}_{f}, \mathbf{x}_o, p)\) and \(\operatorname{dppl}(\mathcal{H}_{l}, \mathcal{H}_{f}, \mathbf{x}_o, p)\) given the length of \(\mathcal{H}_{d}\). We name the task \textbf{\acrfull{dppl-diff}} and the difference \textbf{\acrfull{dppl-diff-metric}}. Higher \acrshort{dppl-diff-metric} value proves that \(\mathcal{H}_{d}\) could predict \(\mathbf{x}_o\) better than \(\mathcal{H}_{l}\). It further indicates that \(\mathcal{H}_{d}\) includes much more useful information about predicting \(\mathbf{x}_o\) than \(\mathcal{H}_{l}\). If one model generally achieves higher \acrshort{dppl-diff-metric}, we regard that it places essential events to \(\mathcal{H}_{d}\), thus is better. The second task asks how many events an approach has to distilled from \(\mathcal{H}_{f}\) to reach given \(\operatorname{dppl}(\mathcal{H}_{d}, \mathcal{H}_{f}, \mathbf{x}_o, p)\) and \(\operatorname{dppl}(\mathcal{H}_{l}, \mathcal{H}_{f}, \mathbf{x}_o, p)\). We call the task \textbf{\acrfull{card-diff}} and the cardinality of \(\mathcal{H}_{d}\) \textbf{\acrfull{card-diff-metric}}. Lower \acrshort{card-diff-metric} means the approach selects events more essential for predicting \(\mathbf{x}_o\), thus is better. We will present the mean and standard deviation(\(\bar{x} \pm \sigma_{x}\))
of \acrshort{dppl-diff-metric} and \acrshort{card-diff-metric} to thoroughly compare \acrshort{model}'s performance with baselines. 

\subsection{Datasets}
\label{sec:datasets}
We train \acrshort{model} on two real-world datasets: Retweet and StackOverflow. During preprocessing, we create the sliding window view of every original event sequence with the window shape equal to the sum of \(\operatorname{card}(\mathcal{H}_{f})\) and \(\operatorname{card}(\mathbf{x}_{o})\). We vary the cardinality of \(\mathcal{H}_{f}\) and \(\mathbf{x}_{o}\) to test how proposed approaches behave under different dataset settings. More details about these datasets are available in \cref{app:model_training}. We discovered that baselines are too slow to evaluate on large datasets(evaluation time longer than 100 hours). To save time, we consider \acrshort{model} and baselines on a smaller subset of the test dataset and only evaluate \acrshort{dppl-diff}/\acrshort{card-diff} performance of \acrshort{model} and \acrshort{dppl-diff} performance of \acrshort{rd} on the complete test datasets. Detailed information about the evaluation speed comparison is available in \cref{sec:speed}.

\section{Experiment Results}

In this section, we present detailed experiment results of \acrshort{ehd} models on real-world datasets. In \cref{sec:proof_l_c_and_L_n}, we report how \acrshort{model} acts if we train it only with \(L_c\) and \(L_n\). We then compare the model performance with baselines in \cref{sec:better_than_random_removal} to prove that our model efficiently selects \(\mathcal{H}_{d}\) that contains rich information for predicting \(\mathbf{x}_o\). Finally, we analyze how \acrshort{model} distills the history via a case study in \cref{sec:case_study}.

\label{sec:results}

\subsection{Effectiveness of \(L_c\) and \(L_n\)}
\label{sec:proof_l_c_and_L_n}
\begin{figure}
    \centering
    \begin{subfigure}[b]{0.4\textwidth}
        \includegraphics[width=\textwidth]{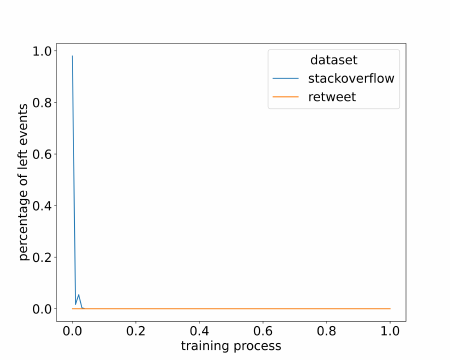}
        \caption{Training \acrshort{model} only with \(L_c\)}
        \label{fig:l_c}
    \end{subfigure}
    \begin{subfigure}[b]{0.4\textwidth}
        \includegraphics[width=\textwidth]{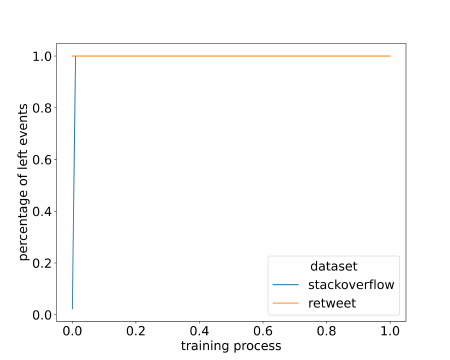}
        \caption{Training \acrshort{model} only with \(L_n\)}
        \label{fig:l_n}
    \end{subfigure}
    \caption{The percentage of left events while training only with \(L_c\) or \(L_n\). Theoretically, optimizing \(L_c\) is supposed to push all events to \(\mathcal{H}_{d}\), causing the percentage of left events to 0\%. Decreasing \(L_n\) does the opposite by moving all events to \(\mathcal{H}_{l}\), thereby lifting the percentage to 100\%. Our experiment results prove that \(L_c\), \(L_n\), and \acrshort{model} work as expected.}
    \label{fig:efficiency}
\end{figure}

In this section, we present experiment results showing the effectiveness of \(L_c\) and \(L_n\). We train the \acrshort{model} with either \(L_c\) or \(L_n\) and show how many percentage of events are left in \(\mathcal{H}_{l}\), \textit{i.e.} \(\operatorname{card}(\mathcal{H}_{l})/\operatorname{card}(\mathcal{H}_{f})\), as the training process goes. Theoretically, solely optimizing \(L_c\) will decrease the percentage to \(0\%\), while optimizing only \(L_n\) will push the percentage to \(100\%\). The experiment results are shown in \cref{fig:efficiency}.

 We could see that in \cref{fig:l_c} when we train the \acrshort{model} with \(L_c\) on all datasets, the percentage of left events quickly hits zero. According to \cref{eqn:L_c}, optimizing \(L_c\) indicates to continuously decrease \(\operatorname{dppl}(\mathcal{H}_{i,l,o,t_l}, \mathcal{H}_{f}, \mathbf{x}_o, p)\). It indicates decreasing \(\log \operatorname{ppl}(p(\mathbf{x}_o|\mathcal{H}_{i,l,o,t_l}))\). The perplexity hits the minimal to \(\log \operatorname{ppl}(p(\mathbf{x}_o|\mathcal{H}_{f}))\) when all \(\mathcal{H}_{i,l,o,t_l}\) is empty, \textit{i.e.} the percentage of left events hits 0\%. In contrast, \cref{eqn:l_n} tells us that \(L_n\) hits the minimal when all \(\hat{y}_i = 0\). This moves all observed historical events to \(\mathcal{H}_{l}\), pushing the percentage of left events to 100\%. The experiment results in \cref{fig:l_n} agree with our theoretical analysis, proving the effectiveness of \(L_c\) and \(L_n\).

\subsection{Comparison between baselines}
\label{sec:better_than_random_removal}
\subsubsection{Evaluation Speed}
\label{sec:speed}
In this section, we compare \acrshort{model}'s evaluation speed with two baselines. The computational complexity of \acrshort{model} and baselines is in \cref{tab:computation_complexity}. We show the computational complexity in \(A/B\) form, where \(A\) is the computational complexity of \acrshort{dppl-diff}, and \(B\) is the computational complexity of \acrshort{card-diff}. We present the average evaluation time ratio between baselines and our approach in \cref{tab:evaluation_time}. These ratios are also presented in \(A/B\) form, where \(A\) and \(B\) refer to the ratio of time a model solving \acrshort{dppl-diff} or \acrshort{card-diff} to \acrshort{model}, respectively. Additionally, we give \(\bar{t}\) the average time of how long all three models finish \acrshort{dppl-diff} and \acrshort{card-diff} on one input sequence.

\cref{tab:computation_complexity} tells that both baselines at least cost linear time in \acrshort{card-diff}: \acrshort{rd} has a linear computational complexity, while \acrshort{gs} has a quadruple one. The reason is both baselines have to guess the length of \(\mathcal{H}_{d}\) by enumeration until they find an appropriate \(\mathcal{H}_{d}\). Baselines have slightly lower computational complexity on \acrshort{dppl-diff}: \acrshort{rd} now obtains one \acrshort{dppl-diff-metric} in constant time because randomly sampling \(M\) \(\mathbf{y}\)s costs constant time. However, \acrshort{gs} still requires quadratic time. \acrshort{model} directly decides which events in \(\mathcal{H}_{f}\) it should distill to \(\mathcal{H}_{d}\) based on \(P(y_i = C|\mathcal{H}_{f}, \mathbf{x}_o)\). This means \acrshort{model} only needs constant time for one \acrshort{dppl-diff-metric} or \acrshort{card-diff-metric}, which is more efficient than baselines.

Evaluation time ratios in \cref{tab:evaluation_time} prove our analysis of computational complexity. \acrshort{gs} achieves last in both evaluation tasks because of its quadruple computational complexity. \acrshort{rd} is consistently faster than \acrshort{gs} on both tasks. Both baselines perform faster on \acrshort{dppl-diff} than \acrshort{card-diff}. However, they are very slow compared with \acrshort{model}. \acrshort{rd} is around five times slower than \acrshort{model} on \acrshort{dppl-diff}, and \acrshort{gs} is over 100 times slower. Conditions are worse on \acrshort{card-diff} where both baselines are 200, even 300 times slower, rendering that \acrshort{model} is the only practical method that scales to large-scale \acrshort{ehd} datasets.

\begin{table}[!ht]
    \centering
    \caption{The computational complexity of our approach and two baselines handling one sequence. \(N\) is the length of \(\mathcal{H}_{f}\), \(M\) is the sample rate of \acrshort{rd}, and \(P\) is the provided length of \(\mathcal{H}_{d}\) while calculating \acrshort{dppl-diff-metric}.}
    \begin{tabular}{lccc}
    \toprule
                                 & \acrshort{model}  &     \acrshort{gs}     & \acrshort{rd}     \\
    \midrule
        Computational Complexity & \(O(1)\)/\(O(1)\) & \(O(P^2)\)/\(O(N^2)\) & \(O(M)\)/\(O(MN)\) \\
    \bottomrule
    \end{tabular}
    \label{tab:computation_complexity}
\end{table}

\begin{table}[!ht]
    \centering
    \caption{The average evaluation time ratio of all \acrshort{ehd} models to \acrshort{model}. A ratio smaller than 1 means this approach is faster than \acrshort{model}, otherwise slower. A bigger ratio means this approach is slower. Ratios are presented in \(A/B\) form: \(A\): speed ratio on \acrshort{dppl-diff}; \(B\): speed ratio on \acrshort{card-diff}. \(\bar{t}\) is the average time evaluating three approaches on one sequence. We obtain these results on the sampled test dataset.}
    \begin{footnotesize}
    \begin{tabular}{lcccccc}
    \toprule
    Dataset & length of \(\mathbf{x}_o\) & length of \(\mathcal{H}_{f}\) & \acrshort{model}(Ours) & \acrshort{gs} & \acrshort{rd} & \(\bar{t}\) \\
    \midrule
        \multirow{5}{*}{Stackoverflow} & 15 & 40 & 1/1 & 122/201 & 5.237/197 & 9.4364s \\
                                       & 15 & 45 & 1/1 & 132/236 & 5.195/213 & 11.536s \\
                                       & 15 & 50 & 1/1 & 150/286 & 5.205/235 & 12.442s \\
                                       & 20 & 50 & 1/1 & 184/310 & 5.199/239 & 13.630s \\
                                       & 25 & 50 & 1/1 & 208/320 & 5.217/241 & 15.285s \\
    \midrule
        \multirow{5}{*}{Retweet} & 10 & 25 & 1/1 & 75.6/101 & 5.668/148 & 6.3702s \\
                                 & 10 & 30 & 1/1 & 98.4/143 & 5.710/173 & 8.3903s \\
                                 & 10 & 35 & 1/1 & 140/204 & 5.732/208 & 10.799s \\
                                 & 15 & 35 & 1/1 & 150/192 & 5.664/204 & 10.877s \\
                                 & 20 & 35 & 1/1 & 160/197 & 5.772/210 & 11.332s \\
    \bottomrule
    \end{tabular}
    \end{footnotesize}
    \label{tab:evaluation_time}
\end{table}

\subsubsection{\acrshort{dppl-diff} Performance}
\label{sec:dppl-diff}
In this section, we present \acrshort{model}'s and baselines \acrshort{dppl-diff} performance. The metric is \acrshort{dppl-diff-metric}, the difference between \(\operatorname{dppl}(\mathcal{H}_{d}, \mathcal{H}_{f}, \mathbf{x}_o, p)\) and \(\operatorname{dppl}(\mathcal{H}_{l}, \mathcal{H}_{f}, \mathbf{x}_o, p)\). It shows how distinctive \(\mathcal{H}_{d}\) and \(\mathcal{H}_{l}\) are regarding predicting \(\mathbf{x}_o\). Because the baselines have efficiency issues, we calculate \acrshort{dppl-diff-metric} of all approaches on a sampled test dataset while only evaluating \acrshort{model} on the full test dataset, which are shown in \cref{tab:ds_mean_and_std_sample} and \cref{tab:ds_mean_and_std_full}.

The results in \cref{tab:ds_mean_and_std_sample} reveal that \(\mathcal{H}_{d}\) and \(\mathcal{H}_{l}\) selected by \acrshort{model} are much distinguishable than other baselines, as we find \acrshort{model}'s \acrshort{dppl-diff-metric} is better than other baselines by a significant margin. That means \acrshort{mtpp} could select essential events more accurately than baselines. All \acrshort{dppl-diff-metric} of \acrshort{rd} is negative, which indicates that its \(\mathcal{H}_{l}\) predictes \(\mathbf{x}_o\) better than \(\mathcal{H}_{d}\). This means \acrshort{rd} consistently fails on picking essential events from \(\mathcal{H}_{f}\). \acrshort{gs} consistently outperforms \acrshort{rd} with its \acrshort{dppl-diff-metric} staying around 0. However, it performs marginally worse than our approach and consumes many computations. 

\begin{table}[!ht]
    \centering
    \caption{The mean and standard deviation of \acrshort{dppl-diff-metric} of \acrshort{model} and baselines. Higher \acrshort{dppl-diff-metric} is better. We obtain these results on the sampled test dataset.}
    \small
    \begin{tabular}{lccccc}
    \toprule
    Dataset & length of \(\mathbf{x}_o\) & length of \(\mathcal{H}_{f}\) & \acrshort{model}(Ours) & \acrshort{gs} & \acrshort{rd} \\
    \midrule
        \multirow{5}{*}{Stackoverflow} & 15 & 40 & \textbf{0.6115\tiny{\(\pm\)0.6477}} & 0.0295\tiny{\(\pm\)1.5293} & -0.5244\tiny{\(\pm\)0.5479} \\
                                       & 15 & 45 & \textbf{0.5716\tiny{\(\pm\)0.7011}} & -0.0761\tiny{\(\pm\)1.5924} & -0.6361\tiny{\(\pm\)0.5893} \\
                                       & 15 & 50 & \textbf{0.5304\tiny{\(\pm\)0.7098}} & -0.1478\tiny{\(\pm\)1.5472} & -0.7266\tiny{\(\pm\)0.6200} \\
                                       & 20 & 50 & \textbf{0.5627\tiny{\(\pm\)0.6020}} & 0.0261\tiny{\(\pm\)2.0056} & -0.5802\tiny{\(\pm\)0.5420} \\
                                       & 25 & 50 & \textbf{0.5850\tiny{\(\pm\)0.5055}} & 0.0579\tiny{\(\pm\)1.4469} & -0.4531\tiny{\(\pm\)0.4696} \\
    \midrule
        \multirow{5}{*}{Retweet} & 10 & 25 & \textbf{0.8138\tiny{\(\pm\)0.3323}} & 0.5355\tiny{\(\pm\)0.4222} & -0.0566\tiny{\(\pm\)0.2065} \\
                                 & 10 & 30 & \textbf{0.8183\tiny{\(\pm\)0.4348}} & 0.5351\tiny{\(\pm\)0.4794} & -0.0877\tiny{\(\pm\)0.2190} \\
                                 & 10 & 35 & \textbf{0.9681\tiny{\(\pm\)0.3798}} & 0.5670\tiny{\(\pm\)0.4744} & -0.0678\tiny{\(\pm\)0.2175} \\
                                 & 15 & 35 & \textbf{0.6617\tiny{\(\pm\)0.2549}} & 0.4357\tiny{\(\pm\)0.3126} & -0.0187\tiny{\(\pm\)0.1561} \\
                                 & 20 & 35 & \textbf{0.5270\tiny{\(\pm\)0.2157}} & 0.3675\tiny{\(\pm\)0.2547} & 0.0103\tiny{\(\pm\)0.1135} \\
    \bottomrule
    \end{tabular}
    \label{tab:ds_mean_and_std_sample}
\end{table}

We further present \acrshort{model} and \acrshort{rd}'s \acrshort{dppl-diff} performances on the complete test dataset in \cref{tab:ds_mean_and_std_full}. We do not evaluate \acrshort{gs} on the full dataset because it is too slow. Nevertheless, we could estimate the \acrshort{dppl-diff} performance of \acrshort{gs} on complete datasets from its results on sampled datasets because other models' performance gap between sampled and full datasets is negligible. Again, \acrshort{model} leads the performance standings, while \acrshort{rd} is still the least. If we borrow \acrshort{gs}'s results in \cref{tab:ds_mean_and_std_sample} as its performance on the complete dataset, it will outperform the Random Search but still marginally worse than \acrshort{model}. These results prove that \acrshort{model} successfully scales to large-scale datasets.

\begin{table}[!ht]
    \centering
    \caption{The mean and standard deviation of \acrshort{dppl-diff-metric} of \acrshort{model} and baselines. Higher \acrshort{dppl-diff-metric} is better. We obtain these results on the full test dataset.}
    \small
    \begin{tabular}{lccccc}
    \toprule
    Dataset & length of \(\mathbf{x}_o\) & length of \(\mathcal{H}_{f}\) & \acrshort{model}(Ours) & \acrshort{gs} & \acrshort{rd} \\
    \midrule
        \multirow{5}{*}{Stackoverflow} & 15 & 40 & \textbf{0.6062\tiny{\(\pm\)0.6685}} & / & -0.5368\tiny{\(\pm\)0.5515} \\
                                       & 15 & 45 & \textbf{0.5667\tiny{\(\pm\)0.6974}} & / & -0.6374\tiny{\(\pm\)0.5890} \\
                                       & 15 & 50 & \textbf{0.5462\tiny{\(\pm\)0.7237}} & / & -0.7290\tiny{\(\pm\)0.6186} \\
                                       & 20 & 50 & \textbf{0.5569\tiny{\(\pm\)0.5983}} & / & -0.5821\tiny{\(\pm\)0.5392} \\
                                       & 25 & 50 & \textbf{0.5768\tiny{\(\pm\)0.5065}} & / & -0.4480\tiny{\(\pm\)0.4707} \\
    \midrule
        \multirow{5}{*}{Retweet} & 10 & 25 & \textbf{0.8192\tiny{\(\pm\)0.3489}} & / & -0.0565\tiny{\(\pm\)0.2115} \\
                                 & 10 & 30 & \textbf{0.8197\tiny{\(\pm\)0.4371}} & / & -0.0893\tiny{\(\pm\)0.2225} \\
                                 & 10 & 35 & \textbf{0.9684\tiny{\(\pm\)0.3773}} & / & -0.0628\tiny{\(\pm\)0.2304} \\
                                 & 15 & 35 & \textbf{0.6571\tiny{\(\pm\)0.2502}} & / & -0.0227\tiny{\(\pm\)0.1591} \\
                                 & 20 & 35 & \textbf{0.5242\tiny{\(\pm\)0.2117}} & / & 0.0079\tiny{\(\pm\)0.1097} \\
    \bottomrule
    \end{tabular}
    \label{tab:ds_mean_and_std_full}
\end{table}

\subsubsection{\acrshort{card-diff} Performance}
In this section, we evaluate \acrshort{model} and other baselines on \acrshort{card-diff} task. This task requires an \acrshort{ehd} model to distill as few events as possible to reach a given \(\operatorname{dppl}(\mathcal{H}_{d}, \mathcal{H}_{f}, \mathbf{x}_o, p)\) and \(\operatorname{dppl}(\mathcal{H}_{l}, \mathcal{H}_{f}, \mathbf{x}_o, p)\). We record and compare the number of distilled events, namely \acrshort{card-diff-metric}, to check which \acrshort{ehd} model could find the shortest \(\mathcal{H}_{d}\) to reach the requirements. The comparison of \acrshort{card-diff-metric} involving all approaches on the sampled test dataset and \acrshort{model} on the full test dataset are presented in \cref{tab:lmhs_mean_and_std_sample} and \cref{tab:lmhs_mean_and_std_full}.

Similar to the conclusions in \cref{sec:dppl-diff}, we could find that \acrshort{model} again significantly outperforms the baselines. Under all conditions, \acrshort{model} achieves the given target by distilling around 50\% fewer events than \acrshort{gs}. This means \acrshort{model} better understands which events are essential for predicting \(\mathbf{x}_o\). \acrshort{gs} owns the second spot, but we could notice that its performance gap to \acrshort{model} is too significant to fill. The gap emphasizes the \acrshort{model}'s superiority to baselines. \acrshort{rd} fails again on \acrshort{card-diff} by achieving the worse performance as it has to select a lot of events from \(\mathcal{H}_{f}\) to reach the requirements. It proves that \acrshort{ehd} is a difficult task worth devising complicated methods to tackle. Moreover, we have to point out that two baselines require a lot of computation for these inferior results, while \acrshort{model} can extract much better \(\mathcal{H}_{d}\) from \(\mathcal{H}_{f}\) in constant time. 

The evaluation time ratios of baselines tell that they are extremely slow on \acrshort{card-diff}. Therefore, we only evaluate \acrshort{model} on the complete test dataset. Again, the performance difference of \acrshort{model} between sampled and full datasets is trivial. This means we can guess baselines' \acrshort{card-diff} performances on complete test sets by results on sampled test sets. The conclusion stays the same as \acrshort{model} still significantly outperforms other baselines. Combined with the evaluation time reports in \cref{sec:speed}, we prove that \acrshort{model} solves \acrshort{ehd} better than baselines in terms of efficiency and the quality of \(\mathcal{H}_{d}\).

\begin{table}[!ht]
    \centering
    \caption{The mean and standard deviation of \acrshort{card-diff-metric} of \acrshort{model} and baselines. Lower \acrshort{card-diff-metric} is better. We obtain these results on the sampled test dataset. }
    \small
    \begin{tabular}{lccccc}
    \toprule
    Dataset & length of \(\mathbf{x}_o\) & length of \(\mathcal{H}_{f}\) & \acrshort{model}(Ours) & \acrshort{gs} & \acrshort{rd} \\
    \midrule
        \multirow{5}{*}{Stackoverflow} & 15 & 40 & \textbf{11.657\tiny{\(\pm\)8.469}} & 22.599\tiny{\(\pm\)12.092} & 35.232\tiny{\(\pm\)7.139} \\
                                       & 15 & 45 & \textbf{11.490\tiny{\(\pm\)9.457}} & 24.401\tiny{\(\pm\)13.948} & 38.982\tiny{\(\pm\)8.348} \\
                                       & 15 & 50 & \textbf{11.333\tiny{\(\pm\)9.941}} & 25.887\tiny{\(\pm\)15.549} & 42.613\tiny{\(\pm\)9.664} \\
                                       & 20 & 50 & \textbf{14.561\tiny{\(\pm\)11.344}} & 28.718\tiny{\(\pm\)15.111} & 43.287\tiny{\(\pm\)9.487} \\
                                       & 25 & 50 & \textbf{16.741\tiny{\(\pm\)11.643}} & 30.214\tiny{\(\pm\)14.783} & 43.889\tiny{\(\pm\)9.109} \\
    \midrule
        \multirow{5}{*}{Retweet} & 10 & 25 & \textbf{10.575\tiny{\(\pm\)3.215}} & 18.173\tiny{\(\pm\)7.194} & 23.950\tiny{\(\pm\)1.769} \\
                                 & 10 & 30 & \textbf{11.254\tiny{\(\pm\)3.988}} & 21.902\tiny{\(\pm\)9.266} & 28.187\tiny{\(\pm\)3.862} \\
                                 & 10 & 35 & \textbf{13.995\tiny{\(\pm\)5.138}} & 27.812\tiny{\(\pm\)9.735} & 34.012\tiny{\(\pm\)1.858} \\
                                 & 15 & 35 & \textbf{15.829\tiny{\(\pm\)5.749}} & 25.792\tiny{\(\pm\)10.519} & 33.885\tiny{\(\pm\)2.045} \\
                                 & 20 & 35 & \textbf{16.517\tiny{\(\pm\)5.298}}  & 25.851\tiny{\(\pm\)10.434} & 34.053\tiny{\(\pm\)1.799} \\
    \bottomrule
    \end{tabular}
    \label{tab:lmhs_mean_and_std_sample}
\end{table}

\begin{table}[!ht]
    \centering
    \caption{The mean and standard deviation of \acrshort{card-diff-metric} of \acrshort{model}(Ours) and baselines. Lower \acrshort{card-diff-metric} is better. We obtain these results on the full test dataset. }
    \small
    \begin{tabular}{lccccc}
    \toprule
    Dataset & length of \(\mathbf{x}_o\) & length of \(\mathcal{H}_{f}\)(Ours) & \acrshort{model} & \acrshort{gs} & \acrshort{rd} \\
    \midrule
        \multirow{5}{*}{Stackoverflow} & 15 & 40 & 11.734\tiny{\(\pm\)8.737} & / & / \\
                                       & 15 & 45 & 11.523\tiny{\(\pm\)9.398} & / & / \\
                                       & 15 & 50 & 11.336\tiny{\(\pm\)9.896} & / & / \\
                                       & 20 & 50 & 14.424\tiny{\(\pm\)11.136} & / & / \\
                                       & 25 & 50 & 16.741\tiny{\(\pm\)11.643} & / & / \\
    \midrule
        \multirow{5}{*}{Retweet} & 10 & 25 & 10.565\tiny{\(\pm\)3.293} & / & / \\
                                 & 10 & 30 & 11.228\tiny{\(\pm\)4.036} & / & / \\
                                 & 10 & 35 & 14.155\tiny{\(\pm\)5.240} & / & / \\
                                 & 15 & 35 & 15.776\tiny{\(\pm\)5.790} & / & / \\
                                 & 20 & 35 & 16.493\tiny{\(\pm\)5.232} & / & / \\
    \bottomrule
    \end{tabular}
    \label{tab:lmhs_mean_and_std_full}
\end{table}

\subsection{Case Studies}
\label{sec:case_study}
In this section, we conduct case studies to reveal how \acrshort{model} distills events from \(\mathcal{H}_{f}\). Firstly, we investigate which mark tends to contain more information about \(\mathbf{x}_o\) by calculating the percentage of each mark getting distilled into \(\mathcal{H}_{d}\). If \acrshort{model} frequently distills events with mark \(i\) to \(\mathcal{H}_{d}\), the percentage goes higher, indicating that abundant historical information aggregates on mark \(i\). Second, we realize that our approach segments \(\mathcal{H}_{f}\) and often moves events from \(\mathcal{H}_{d}\) to \(\mathcal{H}_{l}\) based on these events segments, not on events. Finally, we show the learned \(P(y_i = C|\mathcal{H}_{f}\), \(\mathbf{x}_o)\), and derived \(\mathcal{H}_{d}\) and \(\mathcal{H}_{l}\) on multiple real-world \(\mathcal{H}_{f}\)-\(\mathbf{x}_o)\) pairs. 

\subsubsection{Percentage of distilled events}
\begin{figure}[ht]
    \centering
    \begin{subfigure}{0.45\textwidth}
    \includegraphics[width=\textwidth]{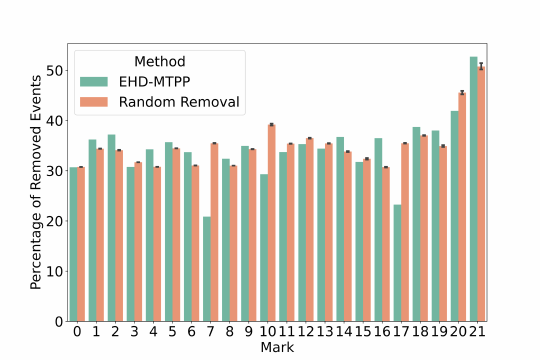}
    \label{fig:percentage_stack_15_40}
    \caption{Stackoverflow(\(\operatorname{len}(\mathbf{x}_{o}) = 15, \operatorname{len}(\mathcal{H}_{f}) = 40\))}
    \end{subfigure}
    \begin{subfigure}{0.45\textwidth}
    \includegraphics[width=\textwidth]{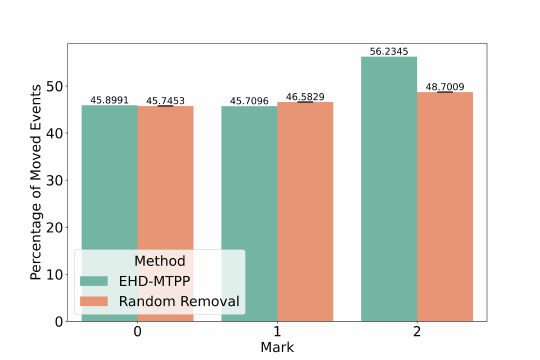}
    \label{fig:percentage_retweet_10_25}
    \caption{Retweet(\(\operatorname{len}(\mathbf{x}_{o}) = 10, \operatorname{len}(\mathcal{H}_{f}) = 25\))}
    \end{subfigure}
    \caption{The percentage of distilled events by \acrshort{model} and \acrshort{rd} on StackOverflow and Retweet dataset. The p-test proves that the distribution of distillation percentages is significantly different from \acrshort{rd}. (I will change the legend name of y to "percentage of distilled events.")}
    \label{fig:percentage}
\end{figure}

Mark \(i\)'s Percentage of distilled events measures the portion of events with mark \(i\) which are distilled to \(\mathcal{H}_{d}\). A higher percentage means \acrshort{model} believes these events help predict the future. Otherwise, the percentage goes lower. We define that \acrshort{model} prefers \(i\)-marked events if the percentage is higher than \acrshort{rd} with high statistical significance. In \cref{fig:percentage}, we exhibit the percentage of each mark getting moved into \(\mathcal{H}_{d}\) by \acrshort{model} and \acrshort{rd}. Results under other dataset hyperparameters are available in (Appendix).

We notice that on StackOverflow, \acrshort{model} dislikes events marked 7, 10, and 17 because the percentage selected by \acrshort{model} is far lower than \acrshort{rd}. For Retweet, the most significant percentage margin between \acrshort{model} and \acrshort{rd} is observed on event mark 2, where \acrshort{model} selects 7.5\% more than \acrshort{rd}. The mark 2 refers to retweets from top-5\% influential users. This means these retweets tend to hold more historical information for prediction, which agrees with the intuition that influential users are more important in information propagation.

\subsubsection{Length Distribution of \(\mathcal{H}_{d}\)}
\label{sec:case_study_length}
\begin{figure}[ht]
    \centering
    \begin{subfigure}{0.49\textwidth}
    \includegraphics[width=\textwidth]{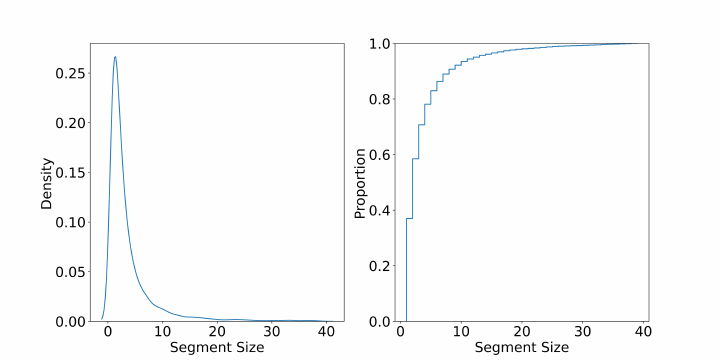}
    \label{fig:length_stack_15_40}
    \caption{Stackoverflow(\(\operatorname{len}(\mathbf{x}_{o}) = 15, \operatorname{len}(\mathcal{H}_{f}) = 40\))}
    \end{subfigure}
    \begin{subfigure}{0.49\textwidth}
    \includegraphics[width=\textwidth]{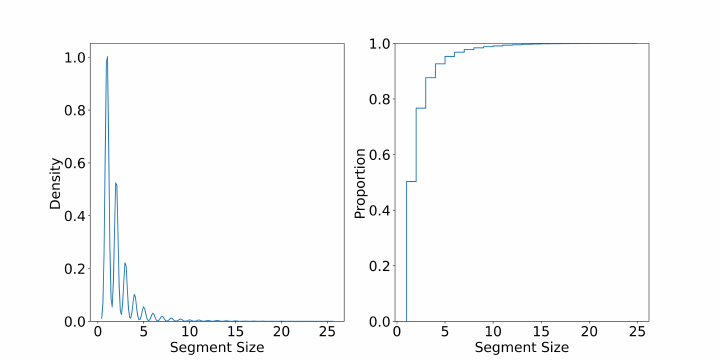}
    \label{fig:length_retweet_10_25}
    \caption{Retweet(\(\operatorname{len}(\mathbf{x}_{o}) = 10, \operatorname{len}(\mathcal{H}_{f}) = 25\))}
    \end{subfigure}
    \caption{The length distribution of \(\mathcal{H}_d\) selected by \acrshort{model} on StackOverflow and Retweet dataset. The distribution of Retweet shows \acrshort{model} favors some lengths. Similar patterns are not observed on StackOverflow.}
    \label{fig:length}
\end{figure}

In this section, we analyze how \acrshort{model} moves historical events from \(\mathcal{H}_{f}\) to \(\mathcal{H}_{d}\). \cref{fig:length} presents the probability density function(PDF, left) and cumulative distribution function(CDF, right) of the length of \(\mathcal{H}_{d}\). We see that the length distribution on StackOverflow is quite smooth. On Retweet, however, the density fluctuates as \(\operatorname{card}(\mathcal{H}_{f})\) increases. It indicates that our model might learn interesting distillation patterns from the Retweet dataset, which we will discuss next.

\begin{figure}[ht]
    \centering
    \includegraphics[width=\textwidth]{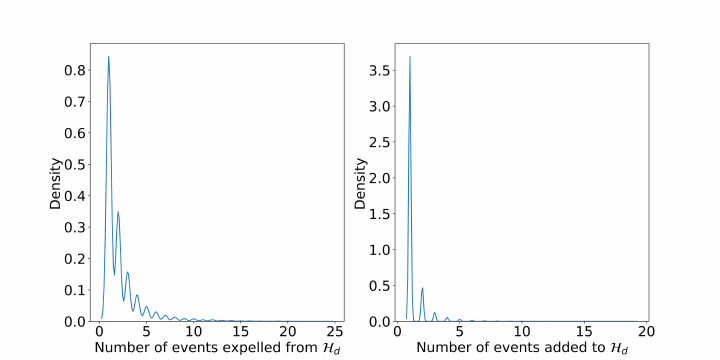}
    \caption{The length distribution of \(\mathcal{H}_d\) selected by \acrshort{model} on Retweet dataset. We could see that \acrshort{model} often adds one event to \(\mathcal{H}_{d}\) as the window sliding on the original event sequence. On the other hand, our approach ejects events out of \(\mathcal{H}_{d}\) by segments.}
    \label{fig:ins_des_length}
\end{figure}

\begin{figure}[ht]
    \centering
    \begin{subfigure}{0.49\textwidth}
    \includegraphics[width=\textwidth]{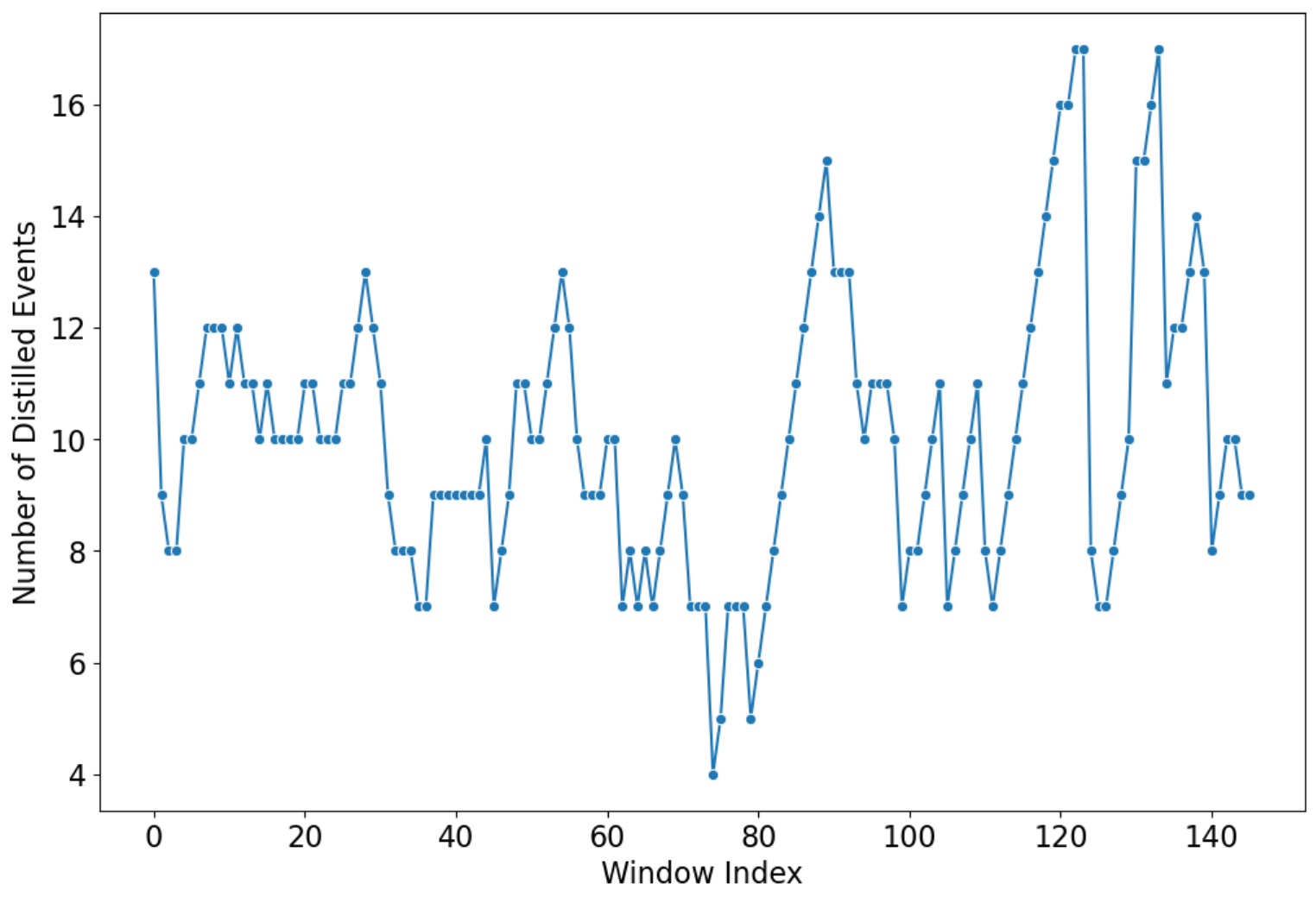}
    \label{fig:length_retweet_example_1}
    \end{subfigure}
    \begin{subfigure}{0.49\textwidth}
    \includegraphics[width=\textwidth]{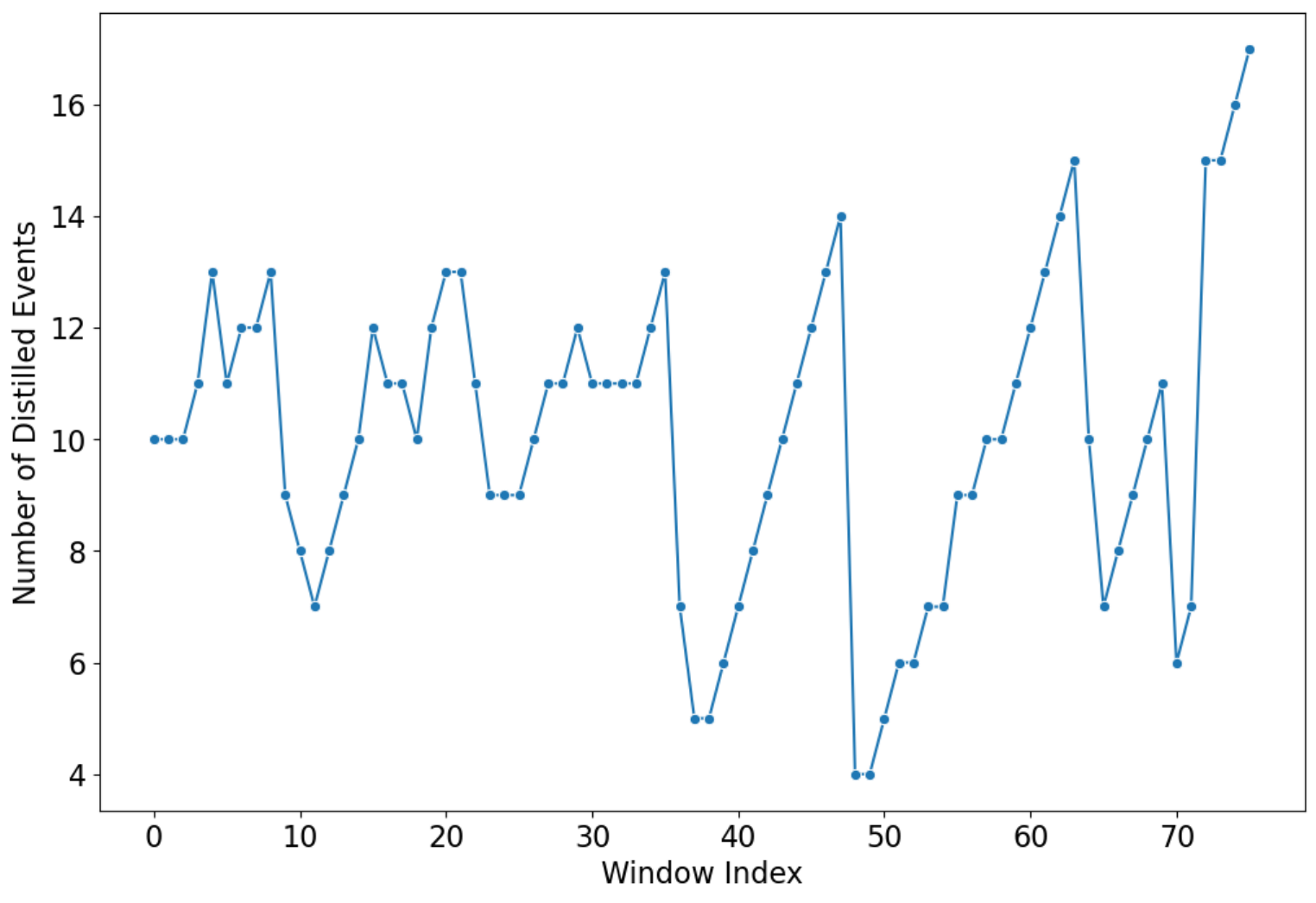}
    \label{fig:length_retweet_example_2}
    \end{subfigure}
    \begin{subfigure}{0.49\textwidth}
    \includegraphics[width=\textwidth]{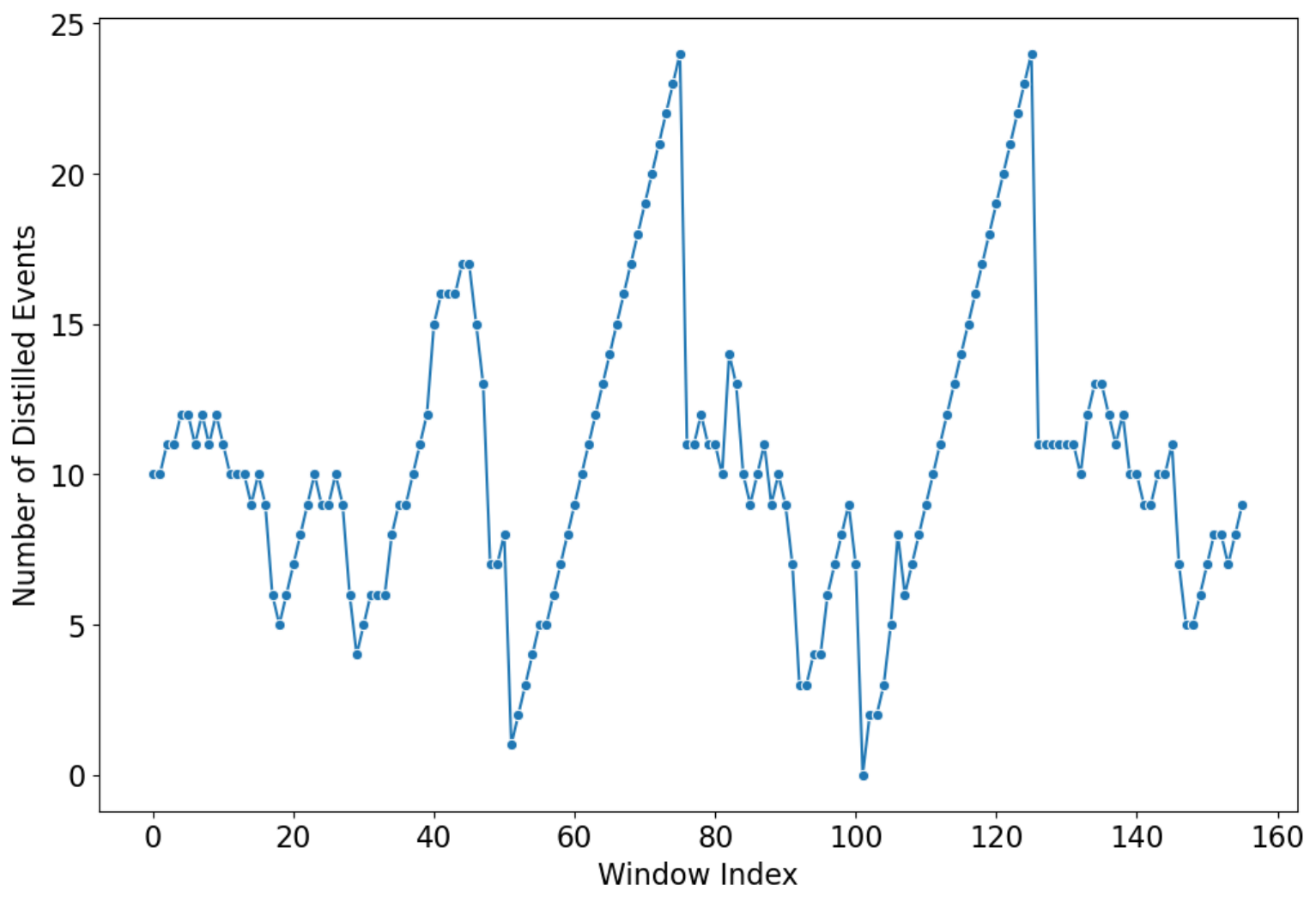}
    \label{fig:length_retweet_example_3}
    \end{subfigure}
    \begin{subfigure}{0.49\textwidth}
    \includegraphics[width=\textwidth]{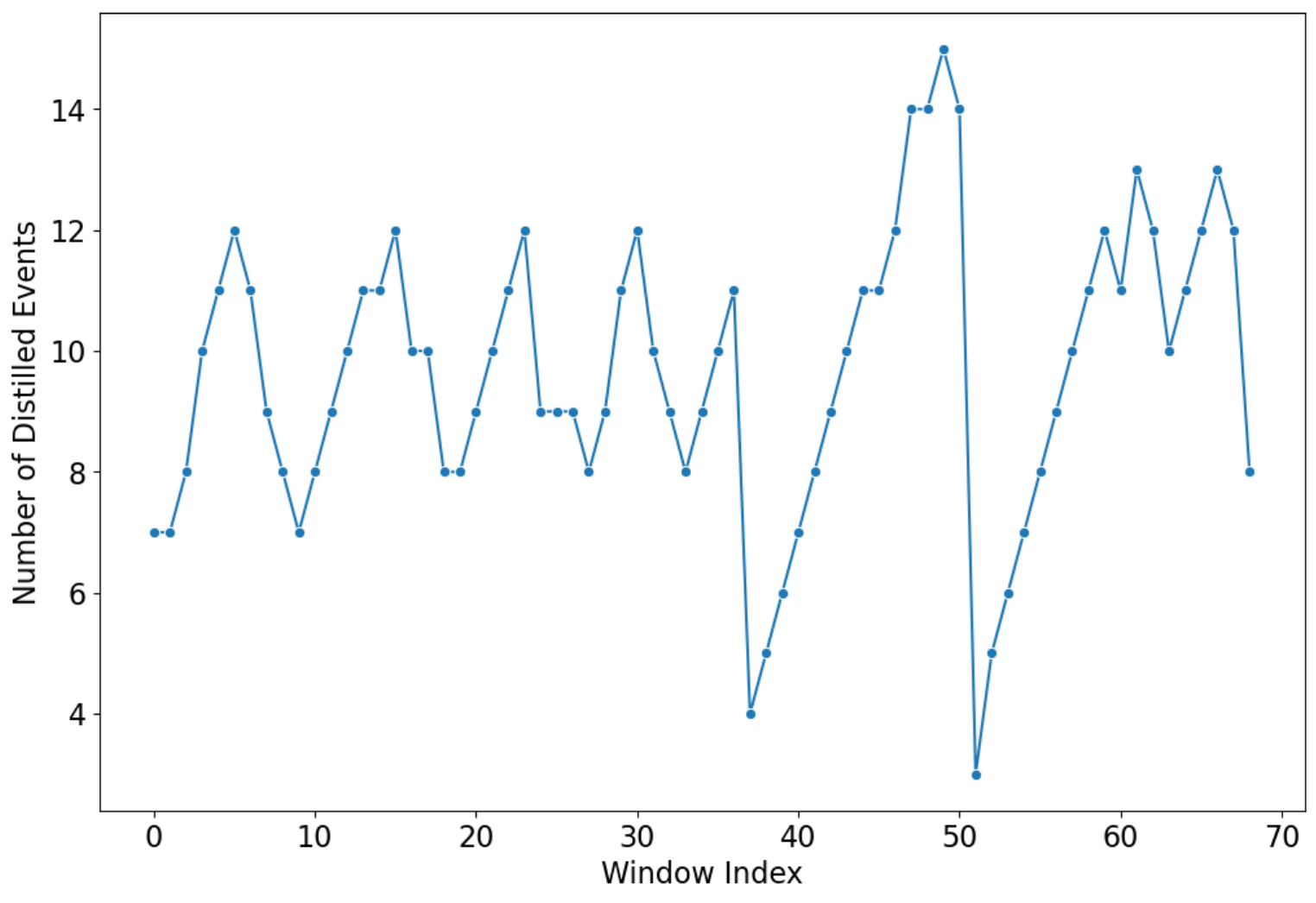}
    \label{fig:length_retweet_example_4}
    \end{subfigure}
    \caption{The length of \(\mathcal{H}_d\) selected by \acrshort{model} on the Retweet dataset. The slow one-by-one increase and drastic drop in the length prove that \acrshort{model} recognizes and distills events in \(\mathcal{H}_{f}\) by segments. We have observed this pattern in almost every sequence in Retweet.}
    \label{fig:length_examples}
\end{figure}

In \cref{sec:datasets} we express that we generate all \(\mathcal{H}_{f}-\mathbf{x}_o\) pairs from the sliding window view of every original event sequence. Given the length of the original sequence as \(l\) and the window size as \(n\), we will extract \(l-n+1\) \(\mathcal{H}_{f}-\mathbf{x}_o\) pairs. For example, if the original sequence has 6 events, and the window size is 3, we will extract \((6-3+1=)\)4 pairs. We want to find out how the length of \(\mathcal{H}_{d}\) varies with the window sliding on the original event sequence. We show the distribution of events added in and ejected from \(\mathcal{H}_d\) in \cref{fig:ins_des_length} and the line plot of \(\operatorname{card}(\mathcal{H}_{d})\) on four original sequences in \cref{fig:length_examples}.

The addition distribution(on the right-hand side) shows that \acrshort{model} often adds events one-by-one into \(\mathcal{H}_d\). Considering we move the sliding window one event each time, this means that \acrshort{model} collects the latest available events into distilled events because they are the temporarily closest events to \(\mathbf{x}_o\). On the other hand, the expel distribution(on the left-hand side) shows that the possibility of moving a lot of events out of \(\mathcal{H}_d\) is significantly higher than adding these into \(\mathcal{H}_d\). It might indicates that \acrshort{model} tends to add events one-by-one to \(\mathcal{H}_{d}\) and move many events from \(\mathcal{H}_{d}\) to \(\mathcal{H}_{l}\) at once. 

To prove our conclusion, we investigate how \acrshort{model} distills events on four sequences. We present how the number of distilled events, \textit{i.e.} \(\operatorname{card}(\mathcal{H}_{d})\), changes w.r.t position of the sliding window in \cref{fig:length_examples}. We see that \acrshort{ehd} often holds historical events into \(\mathcal{H}_{d}\) as \(\operatorname{card}(\mathcal{H}_{d})\) slowly grows, then expels several events as a whole from \(\mathcal{H}_{d}\) as \(\operatorname{card}(\mathcal{H}_{d})\) exhibits a drastic drop. This proves that \acrshort{model} understands that events in \(\mathcal{H}_{f}\) are in groups and successfully recognizes them. 

\section{Limitations and Conclusions}
\textbf{Limitations}
Because most publicly \acrlong{mtpp} datasets \cite{zhao_seismic_2015, Leskovec2014SNAPD, mei_neural_2017, shchur_intensity-free_2020, xue_hypro_2022} lack detailed information about marks and the context of each sequence, we can not translate \(\mathcal{H}_{d}\) into human-understandable languages. This might hurt the explainability of \acrshort{model} for now. Still, we can extract more specific explanations once \acrshort{mtpp} datasets with complete mark information and sequence contexts are available. 

\textbf{Conclusions}
Explainability and accountability of modern machine learning models are essential. In this paper, we show \acrfull{ehd}, a \gls{ca} task to explain \acrshort{mtpp} models by analyzing which events in an observed history sequence are responsible for what has happened recently. We further propose the first machine learning model that tackles \acrshort{ehd}, named \acrfull{model}. \acrshort{model} employs the \acrshort{st-gs} trick to differentially connect inner binary distributions with discrete event masks, which is explicitly required by \acrshort{ehd}'s constraints. Generally, \acrshort{model} provides another view of how to solve discrete optimization problems by machine learning models heuristically. Compared with other baselines, \acrshort{model} performs marginally better on \acrfull{dppl-diff} and \acrfull{card-diff} in terms of efficiency and accuracy.

\clearpage
\bibliography{reference.bib}

\clearpage
\appendix

\section{Model Implementation}\label{app:model_training}
\subsection{The History Rebuilder}
In \cref{sec:solution}, we express how we differentiably convert \(P(y_i|\mathbf{x}_o, \mathcal{H}_{f})\) into \(\hat{\mathbf{y}}\). The central mechanism is the \acrfull{st-gs}. This trick is well-documented and implemented in the PyTorch library as the \(\operatorname{gumbel\_softmax()}\) function with the argument "hard = True." However, here raises two issues. Although \(\hat{\mathbf{y}}\) only contains 0 and 1, this mask tensor is still "continuous" as PyTorch stores \(\hat{\mathbf{y}}\) as a float tensor. PyTorch only accepts selecting items using integers, so we can not use \(\hat{\mathbf{y}}\) to build \(\mathcal{H}_{l}\) or \(\mathcal{H}_{d}\). We also can not convert \(\hat{\mathbf{y}}\) to an integer tensor \(\mathbf{y}\) because such conversion inevitably loses \(\hat{\mathbf{y}}\)'s gradient. The other issue is selecting items from a sequence also introduces the gradient of the original sequence, not the index tensor. For example, we have a mask tensor \(\mathbf{y} = (1, 0)\) with gradient \(g(\mathbf{y}) = (g(y_1), g(y_2))\) and an event sequence \(\mathbf{e} = (e_1, e_2)\) with gradient \(g(\mathbf{e}) = (g(e_1), g(e_2))\). \(1\) in the mask tensor refers to acceptance while \(0\) refers to rejection. This means the selected sequence \(\mathbf{e}_{l} = (e_1)\). However, the gradient of \(\mathbf{e}_{l}\) is \(g(e_1)\), while \cref{fig:ehd_model} indicates that \(\mathbf{e}_{l}\)'s gradient should be \(g(y_1)\) because only the gradient coming from \(\hat{\mathbf{y}}\) optimizes \acrshort{model}.

To tackle these issues, we attach the gradient of \(\mathbf{y}\) to the event sequence \(\mathbf{e}\) by matrix multiplication. First, we multiply \(\hat{\mathbf{y}}\) with all continuous representations, \textit{i.e.} event embeddings \(\mathbf{E}\) and time intervals \(\mathbf{T}\). This attaches the gradient of \(\hat{\mathbf{y}}\) to \(\mathbf{E}\) and \(\mathbf{T}\). We denote them as \(\hat{\mathbf{E}}\) and \(\hat{\mathbf{T}}\), respectively. Next, we use the converted \(\mathbf{y}\) to select representations of \(\mathcal{H}_{l}\) or \(\mathcal{H}_{d}\) from \(\hat{\mathbf{E}}\) and \(\hat{\mathbf{T}}\), and other discrete features, such as events and masks. Here we present the code of the function "filter()" to show how we rebuild \(\mathcal{H}_{l}\) or \(\mathcal{H}_{d}\) from \(\hat{\mathbf{y}}\)\footnote{We remove some auxiliary code for the sake of space saving. This means the presented function could differ from the source code's real "filter()" function. However, the differences should not affect the final result.}.

\begin{python}
def filter(self, input_time, input_events, events_embeddings, input_mask, filter_mask):
    samples_for_l_p, batch_size \
        = filter_mask.shape[0], filter_mask.shape[1]

    # We remove the event whose \hat{y}_i == 1.
    # This means all left events have their \hat{y}_i = 0
    filter_mask_for_nominated = filter_mask[..., 0]
    discrete_filter_mask_for_nominated = filter_mask[..., 0].detach().int()

    the_number_of_remained_event = discrete_filter_mask_for_nominated.sum(dim = -1)
    [repeated_input_time, repeated_input_events, \
    repeated_events_embeddings, repeated_input_mask] \
    = repeat([repeated_input_time, repeated_input_events, \
              repeated_events_embeddings, repeated_input_mask] , \
              '... -> n ...', n = samples_for_l_p)
    repeated_cumsum_time = repeated_input_time.cumsum(dim = -1)
        
    # select the remaining events from the original input.
    # where the gradient attachment happens.
    selected_time = repeated_cumsum_time * filter_mask_for_nominated
    selected_time = selected_time[discrete_filter_mask_for_nominated == 1]
    selected_input_events \
        = repeated_input_events[discrete_filter_mask_for_nominated == 1]
    selected_events_embeddings \
        = repeated_events_embeddings * filter_mask_for_nominated.unsqueeze(dim = -1)
    selected_events_embeddings \
        = selected_events_embeddings[discrete_filter_mask_for_nominated == 1]
    selected_input_mask \
        = repeated_input_mask[discrete_filter_mask_for_nominated == 1]
        
    data_start_index = 0
    all_reshaped_time, all_reshaped_input_events, \
    all_reshaped_events_embeddings, all_reshaped_input_mask \
        = [], [], [], []
    for the_number_of_remained_event_per_batch \
        in the_number_of_remained_event:
        # Padding the left sequences.
        reshaped_time, reshaped_input_events, reshaped_events_embeddings,
            reshaped_input_mask = [], [], [], []
        for the_number_of_remained_event_per_batch_per_seq \
            in the_number_of_remained_event_per_batch:
            reshaped_time.append(\
            selected_time[data_start_index:data_start_index +\
                the_number_of_remained_event_per_batch_per_seq])
            reshaped_input_events.append(\
            selected_input_events[data_start_index:data_start_index +\
                the_number_of_remained_event_per_batch_per_seq])
            reshaped_events_embeddings.append(\
            selected_events_embeddings[data_start_index:data_start_index +\
            the_number_of_remained_event_per_batch_per_seq, :])
            reshaped_input_mask.append(\
            selected_input_mask[data_start_index:data_start_index +\
            the_number_of_remained_event_per_batch_per_seq])
            
            data_start_index += \
                the_number_of_remained_event_per_batch_per_seq
                        
        padded_reshaped_time \
            = torch.nn.utils.rnn.pad_sequence(reshaped_time, batch_first = True)
        padded_input_events \
            = torch.nn.utils.rnn.pad_sequence(reshaped_input_events, batch_first = True)
        padded_events_embeddings \
            = torch.nn.utils.rnn.pad_sequence(reshaped_events_embeddings, batch_first = True)
        padded_input_mask \
            = torch.nn.utils.rnn.pad_sequence(reshaped_input_mask, batch_first = True)            
        padded_reshaped_time \
        = padded_reshaped_time.diff(dim = -1, \
          prepend = torch.zeros(batch_size, 1, device = self.device))

        all_reshaped_time.append(padded_reshaped_time)
        all_reshaped_input_events.append(padded_input_events)
        all_reshaped_events_embeddings.append(padded_events_embeddings)
        all_reshaped_input_mask.append(padded_input_mask)

        return all_reshaped_time, all_reshaped_input_events, \
               all_reshaped_events_embeddings, all_reshaped_input_mask
\end{python}

\subsection{Datasets}
In this paper, we evaluate \acrshort{model} and other baselines on two datasets: retweet and StackOverflow. Detailed statistical features of these datasets are listed in \cref{tab:dataset_features}. 


\textbf{Retweet}\cite{zhao_seismic_2015} records when users retweet a particular message on Twitter. The mark of this dataset distinguishes all users into three different types: (1) normal user, whose followers count is lower than the median, (2) influence user, whose followers count is higher than the median but lower than the $95$th percentile, (3) famous user, whose followers count is higher than the $95$th percentile. 

\textbf{StackOverflow}\cite{Leskovec2014SNAPD} was collected from Stackoverflow\footnote{https://stackoverflow.com/}, a popular question-answering website about various topics. Users providing decent answers will receive different badges as rewards. 

\begin{table}[!ht]
    \centering
    \small
    \caption{Statistical features of used four real-world datasets. The number of sequences and events aggregates the training, evaluation, and test datasets. \(\bar{\tau}\) and \(\sigma(\tau)\) are the mean and standard deviation of the time interval between two adjacent events. \(t_0\) and \(T\) are an event sequence's start and end times.}
    \begin{tabular}{lccccccc}
        \toprule
                      & Number of sequences & Number of events & Number of marks & \(\bar{\tau}\) & \(\sigma(\tau)\)  & \(t_0\) & \(T\)   \\
        \midrule
        Retweet       & 24,000 & 2,610,102 & 3 & 2,574  & 16,302 & 0    & 604,799  \\
        StackOverflow & 6,633  & 480,414  & 22 & 0.8747 & 1.2091 & 1,324 & 1,390    \\
        \bottomrule
    \end{tabular}
    \label{tab:dataset_features}
\end{table}

\subsection{Hyperparameters}
\subsubsection{Datasets}
In this section, we present the hyperparameters used to generate datasets for \acrshort{model} from the original datasets in \cref{tab:hyperparameter_dataset} and the size of training, evaluation, full test, and sampled test datasets in \cref{tab:dataset_size}. We generate datasets by sliding window. The size of the window is the sum of \(\operatorname{len}(\mathbf{x}_o)\) and \(\operatorname{len}(\mathcal{H}_{f})\), which are the length of \(\mathbf{x}_o\) and \(\mathcal{H}_{f}\), respectively. 

\begin{table}[!ht]
    \centering
    \caption{Hyperparameters used to generate datasets for \acrshort{model}}
    \begin{tabular}{lc}
       \toprule
                     &  Available \(\operatorname{len}(\mathbf{x}_o)\)-\(\operatorname{len}(\mathcal{H}_{f})\) pairs \\
       \midrule
       Retweet       &          (10, 25), (10, 30), (10, 35), (15, 35), (20, 35)               \\
       StackOverflow &          (15, 40), (15, 45), (15, 50), (20, 50), (25, 50)               \\
       \bottomrule
    \end{tabular}
    \label{tab:hyperparameter_dataset}
\end{table}

\begin{table}[!ht]
    \centering
    \caption{Properties of all generated \acrshort{ehd} datasets}
    \begin{tabular}{lccccc}
       \toprule
                     &  \(\operatorname{len}(\mathbf{x}_o)\)-\(\operatorname{len}(\mathcal{H}_{f})\) pair & len(training) & len(evaluation) & len(test\_full) & len(test\_sampled) \\
       \midrule
       \multirow{5}{*}{Retweet} & (10, 25)  & 1,476,116 & 145,521 & 148,465 & 5,000 \\
                                & (10, 30)  & 1,376,116 & 135,521 & 135,521 & 5,000 \\
                                & (10, 35)  & 1,276,116 & 125,521 & 128,465 & 5,000 \\
                                & (15, 35)  & 1,176,383 & 115,551 & 118,497 & 5,000 \\
                                & (20, 35)  & 1,081,289 & 106,047 & 108,970 & 5,000 \\
        \midrule
       \multirow{5}{*}{StackOverflow} & (15, 40) & 99,791 & 10,826 & 29,232 & 5,000 \\
                                      & (15, 45) & 87,623 & 9,451 & 25,824 & 5,000 \\
                                      & (15, 50) & 77,341 & 8,307 & 22,951 & 5,000 \\
                                      & (20, 50) & 68,635 & 7,350 & 20,504 & 5,000 \\
                                      & (25, 50) & 61,254 & 6,512 & 18,385 & 5,000 \\
       \bottomrule
    \end{tabular}
    \label{tab:dataset_size}
\end{table}

\subsubsection{Models}
Theoretically, \acrshort{model} can use any MTPP models that can provide \(p^*(m, t)\). In this paper, we use FullyNN\cite{omi_fully_2019} to calculate \(p^*(m, t)\) for \acrshort{model}. \cref{tab:fullynn_hyperparameters} and \cref{tab:ehd_hyperparameters} present the hyperparameter used for training the FullyNN and \acrshort{model} on Retweet and StackOverflow, respectively. 

FullyNN directly estimates the integral of intensity functions \(\Lambda^*(t) = \int_{t_l}^{t}{\lambda^*(\tau)d\tau}\), also known as the compensator, and calculates the value of the intensity function at \(t\) from the gradient of \(\Lambda^*(t)\), \textit{i.e.}:
\begin{align}
    \Lambda^*(t) &= \int_{t_l}^{t}{\lambda^*(\tau)d\tau} = \operatorname{FullyNN}(t) \\
    \lambda^*(t) &= \frac{\partial \Lambda^*(t)}{\partial t} = \frac{\partial \operatorname{FullyNN}(t)}{\partial t} \\
    p^*(t) = \lambda^*(t)\exp(-\Lambda^*(t)) &= \frac{\partial \operatorname{FullyNN}(t)}{\partial t}\exp(-\operatorname{FullyNN}(t))
\end{align}
This helps FullyNN elude calculating \(\Lambda^*(t)\) by numerical integration methods, such as Monte Carlo integration, to predict \acrshort{mtpp} faster and more accurately. However, Shchur et al.\cite{shchur_intensity-free_2020} point out that FullyNN learns a malformed distribution, which might lead to poor extrapolation performances and wrong time predictions.

\begin{table}[!ht]
    \centering
    \caption{Hyperparamters settings for the MTPP model}
    \begin{tabular}{lcc}
    \toprule
                     & Retweet  & StackOverflow \\
    \midrule
    Training Steps   & 400,000  & 200,000  \\
    Warmup Steps     & 80,000   & 40,000   \\
    Batch Size       & 32       & 32       \\
    History Embedding& 32       & 32       \\
    Intensity Vector & 16       & 32       \\
    Learning Rate    & 0.002    & 0.002    \\
    Layers           & 4        & 2        \\
    \bottomrule
    \end{tabular}
    \label{tab:fullynn_hyperparameters}
\end{table}

\begin{table}[!ht]
    \centering
    \caption{Caption}
    \begin{tabular}{lcc}
    \toprule
                      & Retweet  & StackOverflow \\
    \midrule
        Training Steps& 100,000  & 100,000 \\
        Warmup Steps  & 2,000    & 2,000   \\
        Batch Size    & 256      & 128     \\
        Hidden Vector & 64       & 64      \\
        Input Vector  & 32       & 32      \\
          Q, K, V     & 32       & 32      \\
          Head        & 4        & 4       \\
          N           & 4        & 4       \\
          M           & 4        & 4       \\
        Learning Rate & 0.001    & 0.001   \\
        \(\epsilon\)  & 0.5      & 0.5     \\
        \(\alpha\)    & 1.0      & 1.0     \\
    \bottomrule
    \end{tabular}
    \label{tab:ehd_hyperparameters}
\end{table}

\section{Baseline Models}
\label{app:baselines}
In the main paper, we have mentioned two \acrshort{ehd} baseline models: \textbf{\acrshort{rd}} and \textbf{\acrshort{gs}}. Here, we give further details, including algorithms, about how these two baselines handle \acrshort{dppl-diff} and \acrshort{card-diff}. 

\subsection{Solving \acrshort{dppl-diff}}
In \cref{sec:solution}, we express that \acrshort{model} solves a relaxed \acrshort{ehd} question where \(\operatorname{dppl}(\mathcal{H}_{l}, \mathcal{H}_{f}, \mathbf{x}_o, p)\) should not be larger than \(\log \epsilon\). As a result, \(\operatorname{dppl}(\mathcal{H}_{l}, \mathcal{H}_{f}, \mathbf{x}_o, p)\) provided by \acrshort{ehd} is always smaller than \(\log \epsilon\). This means that baselines can not use \(\epsilon\) to search for \(\mathcal{H}_{l}\) and \(\mathcal{H}_{d}\), otherwise the results of \acrshort{model} and baselines are not comparable. We instead consider both \(\log \operatorname{ppl}(p(\mathbf{x}_o|\mathcal{H}_{d}))\) and \(\log \operatorname{ppl}(p(\mathbf{x}_o|\mathcal{H}_{l}))\). Specifically, \acrshort{model} gives out its \(dppl_{r, model} = \operatorname{dppl}(\mathcal{H}_{d}, \mathcal{H}_{f}, \mathbf{x}_o, p)\) and \(dppl_{l, model} = \operatorname{dppl}(\mathcal{H}_{l}, \mathcal{H}_{f}, \mathbf{x}_o, p)\). We ask all baselines to find their minimal \(\mathcal{H}_{d}\) so that \(dppl_{r, baseline} > dppl_{r, model}\) and \(dppl_{l, baseline} < dppl_{l, model}\). In other words, we want to check which approach could reach a known prediction performance by selecting the fewest events. The model that can achieve the requirement with the smallest \(\mathcal{H}_{d}\) is considered the best. Here, we present the algorithm of two baselines in \cref{alg:random_given_ppl} and \cref{alg:greedy_given_ppl} under this setting.

\begin{algorithm}[tb]
\caption{Decide \(\mathcal{H}_{d}\) from \(\mathcal{H}_{f}\) by \acrshort{rd} with known \(dppl_{r, model}\) and \(dppl_{l, model}\).}
\label{alg:random_given_ppl}
\begin{algorithmic}
\STATE {\bfseries Input:} The observed history \(\mathcal{H}_{f}\), observed short future \(\mathbf{x}_o\), target performance \(dppl_{r, model}\) and \(dppl_{l, model}\), sample rate \(M\);
\STATE {\bfseries Output:} the number of selected events;
\STATE \(\mathcal{H}_{d} = \Phi, num\_of\_selected\_item = 0\)
\STATE \(dppl_r = \operatorname{dppl}(\mathcal{H}_{d}, \mathcal{H}_{f}, \mathbf{x}_o, p_{model}), dppl_l = \operatorname{dppl}(\mathcal{H}_{l}, \mathcal{H}_{f}, \mathbf{x}_o, p_{model})\)
\WHILE{\(dppl_r < dppl_{r, model}\) or \(dppl_l > dppl_{l, model}\)}
\STATE \(num\_of\_selected\_item = num\_of\_selected\_item + 1\)
\STATE \(dppl\_r\_list = \operatorname{list}(), dppl\_l\_list = \operatorname{list}()\)
\FOR {\(index\) in range(\(M\))}
\STATE randomly sample a \(\mathbf{y}\) where \(\sum_{y_i \in \mathbf{y}}{y_i} = num\_of\_selected\_item\).
\STATE build \(\mathcal{H}_{l}\) and \(\mathcal{H}_{d}\) from the sampled \(\mathbf{y}\).
\STATE \(dppl\_r\_list\).push(\(\operatorname{dppl}(\mathcal{H}_{d}, \mathcal{H}_{f}, \mathbf{x}_o, p_{model})\))
\STATE \(dppl\_l\_list\).push(\(\operatorname{dppl}(\mathcal{H}_{l}, \mathcal{H}_{f}, \mathbf{x}_o, p_{model})\))
\ENDFOR
\STATE \(dppl_r = dppl\_r\_list\).mean()
\STATE \(dppl_l = dppl\_l\_list\).mean()
\ENDWHILE
\RETURN \(num\_of\_selected\_item\)
\end{algorithmic}
\end{algorithm}

\begin{algorithm}[tb]
\caption{Decide \(\mathcal{H}_{d}\) from \(\mathcal{H}_{f}\) by \acrshort{gs} with known \(dppl_{r, model}\) and \(dppl_{l, model}\).}
\label{alg:greedy_given_ppl}
\begin{algorithmic}
\STATE {\bfseries Input:} The observed history \(\mathcal{H}_{f}\), observed short future \(\mathbf{x}_o\), target performance \(dppl_{r, model}\) and \(dppl_{l, model}\);
\STATE {\bfseries Output:} the number of selected events;
\STATE \(\mathcal{H}_{d} = \Phi, num\_of\_selected\_item = 0\)
\STATE \(dppl_r = \operatorname{dppl}(\mathcal{H}_{d}, \mathcal{H}_{f}, \mathbf{x}_o, p_{model}), dppl_l = \operatorname{dppl}(\mathcal{H}_{l}, \mathcal{H}_{f}, \mathbf{x}_o, p_{model})\)
\WHILE{\(dppl_r < dppl_{r, model}\) or \(dppl_l > dppl_{l, model}\)}
\STATE \(h\_list = \operatorname{list}()\)
\STATE \(dppl\_r\_list = \operatorname{list}(), dppl\_l\_list = \operatorname{list}()\)
\FOR {\( x_i \in \mathcal{H}_{f} \)}
\IF {\(x_i \in \mathcal{H}_{d} \)}
\STATE continue
\ENDIF
\STATE \(\mathcal{H}_{tmp, r, o, t_l} = x_i \cup \mathcal{H}_{d}\)
\STATE \(dppl\_r\_list\).push(\(\operatorname{dppl}(\mathcal{H}_{tmp, r, o, t_l}, \mathcal{H}_{f}, \mathbf{x}_o, p_{model})\))
\STATE \(dppl\_l\_list\).push(\(\operatorname{dppl}(\mathcal{H}_{f} - \mathcal{H}_{tmp, r, o, t_l}, \mathcal{H}_{f}, \mathbf{x}_o, p_{model})\))
\STATE \(h\_list\).push(\(\mathcal{H}_{tmp, r, o, t_l}\))
\ENDFOR
\STATE \(dppl_l = \min(ppl\_l\_list)\)
\STATE \(dppl_r = dppl\_r\_list[\operatorname{indexof}(dppl_l)]\)
\STATE \(\mathcal{H}_{d} = h\_list[\operatorname{indexof}(dppl_l)]\)
\STATE \(num\_of\_selected\_item = num\_of\_selected\_item + 1\)
\ENDWHILE
\RETURN \(num\_of\_selected\_item\)
\end{algorithmic}
\end{algorithm}

\subsection{Solving \acrshort{card-diff}}
Another way to rank \acrshort{ehd} approaches examines how much useful information \(\mathcal{H}_{d}\) selected by each approach contains for predicting \(\mathbf{x}_o\) given the length of \(\mathcal{H}_{d}\). Specifically, all approaches know that they must select \(l_r\) events from \(\mathcal{H}_{f}\), and their target is to minimize \(\operatorname{dppl}(\mathcal{H}_{l}, \mathcal{H}_{f}, \mathbf{x}_o, p_{model})\). \acrshort{model} intrinsically judge how many events should be distilled. We can not force it to select a given number of events from \(\mathcal{H}_{f}\). Therefore, we pick the length of \(\mathcal{H}_{d}\) generated by \acrshort{model} as \(l_r\) and feed it to the baselines which accepts predefined \(l_r\). We present the algorithm of two baselines in \cref{alg:random_given_n} and \cref{alg:greedy_given_n} under the fixed-\(l_r\) setting.

\begin{algorithm}[tb]
\caption{Decide \(\mathcal{H}_{d}\) from \(\mathcal{H}_{f}\) by \acrshort{rd} with predefined \(l_d\).}
\label{alg:random_given_n}
\begin{algorithmic}
\STATE {\bfseries Input:} The observed history \(\mathcal{H}_{f}\), observed short future \(\mathbf{x}_o\), the number of distilled events \(l_d\), sample rate \(M\);
\STATE {\bfseries Output:} \(\operatorname{dppl}(\mathcal{H}_{l}, \mathcal{H}_{f}, \mathbf{x}_o, p_{model})\) and \(\operatorname{dppl}(\mathcal{H}_{d}, \mathcal{H}_{f}, \mathbf{x}_o, p_{model})\);
\STATE randomly sample \(M\) \(\mathbf{y}\)s where \(\sum_{y_i \in \mathbf{y}}{y_i} = l_d\).
\STATE build \(M\) \(\mathcal{H}_{l}\)-\(\mathcal{H}_{d}\) pairs from the sampled \(\mathbf{y}\)s.
\STATE \(dppl\_r\_list = \operatorname{dppl}(\mathcal{H}_{d}, \mathcal{H}_{f}, \mathbf{x}_o, p_{model})\))
\STATE \(dppl\_l\_list = \operatorname{dppl}(\mathcal{H}_{l}, \mathcal{H}_{f}, \mathbf{x}_o, p_{model})\))
\RETURN \(dppl\_r\_list\).mean(), \(dppl\_l\_list\).mean()
\end{algorithmic}
\end{algorithm}

\begin{algorithm}[tb]
\caption{Decide \(\mathcal{H}_{d}\) from \(\mathcal{H}_{f}\) by \acrshort{gs} with predefined \(l_d\).}
\label{alg:greedy_given_n}
\begin{algorithmic}
\STATE {\bfseries Input:} The observed history \(\mathcal{H}_{f}\), observed short future \(\mathbf{x}_o\), the number of distilled events \(l_d\);
\STATE {\bfseries Output:} \(\operatorname{dppl}(\mathcal{H}_{l}, \mathcal{H}_{f}, \mathbf{x}_o, p_{model})\) and \(\operatorname{dppl}(\mathcal{H}_{d}, \mathcal{H}_{f}, \mathbf{x}_o, p_{model})\);
\STATE \(\mathcal{H}_{d} = \Phi, num\_of\_removed\_item = 0\)
\STATE \(dppl_r = \operatorname{dppl}(\mathcal{H}_{d}, \mathcal{H}_{f}, \mathbf{x}_o, p_{model}), dppl_l = \operatorname{dppl}(\mathcal{H}_{l}, \mathcal{H}_{f}, \mathbf{x}_o, p_{model})\)
\WHILE{\( num\_of\_removed\_item \neq l_d\)}
\STATE \(h\_list = \operatorname{list}()\)
\STATE \(dppl\_r\_list = \operatorname{list}(), dppl\_l\_list = \operatorname{list}()\)
\FOR {\( x_i \in \mathcal{H}_{f} \)}
\IF {\(x_i \in \mathcal{H}_{d} \)}
\STATE continue
\ENDIF
\STATE \(\mathcal{H}_{tmp, r, o, t_l} = x_i \cup \mathcal{H}_{d}\)
\STATE \(dppl\_r\_list\).push(\(\operatorname{dppl}(\mathcal{H}_{tmp, r, o, t_l}, \mathcal{H}_{f}, \mathbf{x}_o, p_{model})\))
\STATE \(dppl\_l\_list\).push(\(\operatorname{dppl}(\mathcal{H}_{f} - \mathcal{H}_{tmp, r, o, t_l}, \mathcal{H}_{f}, \mathbf{x}_o, p_{model})\))
\STATE \(h\_list\).push(\(\mathcal{H}_{tmp, r, o, t_l}\))
\ENDFOR
\STATE \(dppl_l = \min(ppl\_l\_list)\)
\STATE \(dppl_r = dppl\_r\_list[\operatorname{indexof}(dppl_l)]\)
\STATE \(\mathcal{H}_{d} = h\_list[\operatorname{indexof}(dppl_l)]\)
\STATE \(num\_of\_removed\_item = num\_of\_removed\_item + 1\)
\ENDWHILE
\RETURN \(dppl_r\), \(dppl_l\)
\end{algorithmic}
\end{algorithm}

\printglossary
\end{document}